%% file: paper.tex
\documentclass[10pt,twocolumn,letterpaper]{article}

\usepackage{iccv}
\usepackage{times}
\usepackage{epsfig}
\usepackage{graphicx}
\usepackage{amsmath}
\usepackage{amssymb}

\usepackage{caption}
\usepackage{subcaption}
\usepackage{epsfig}
\usepackage{booktabs}
\usepackage{rotating}
\usepackage{algorithm} % algorithm
\usepackage{algpseudocode}
\usepackage{multirow}
\usepackage{color}

\usepackage{enumitem}
\usepackage{enumerate}

\newif\ifshowcomments
\showcommentstrue
\ifshowcomments
\newcommand{\TODO}[1]{{\color{red}{[TODO: #1]}}}
\newcommand{\revised}[1]{{\color[rgb]{0.2,0.7,0.2}{#1}}}
\newcommand{\ytjiang}[1]{{\color[rgb]{0.7,0.7,0}{[YT: #1]}}}
\newcommand{\phil}[1]{{\color[rgb]{0.9,0.1,0.1}{[#1]}}}

\else
\newcommand{\TODO}[1]{}
\newcommand{\revised}[1]{}
\newcommand{\ytjiang}[1]{}
\newcommand{\phil}[1]{}
\fi

\usepackage[pagebackref=true,breaklinks=true,letterpaper=true,colorlinks,bookmarks=false]{hyperref}

\iccvfinalcopy % *** Uncomment this line for the final submission

\def\iccvPaperID{2609} % *** Enter the ICCV Paper ID here

\ificcvfinal\fi
\begin{document}
	
	%%%%%%%%% TITLE
	\title{Mask-ShadowGAN: Learning to Remove Shadows from Unpaired Data}
	
	\author{Xiaowei Hu$^{1}$,  Yitong Jiang$^{2}$, Chi-Wing Fu$^{1,2,\ast}$, and Pheng-Ann Heng$^{1,2,}\thanks{Co-corresponding authors}$\\
		$^1$ Department of Computer Science and Engineering, The Chinese University of Hong Kong\\
		$^2$ Guangdong Provincial Key Laboratory of Computer Vision and Virtual Reality Technology, \\ Shenzhen Institutes of Advanced Technology, Chinese Academy of Sciences, China}
	
\maketitle
\thispagestyle{empty}

%%%%%%%%% ABSTRACT
\begin{abstract}
%Shadow removal is a fundamental and challenging task, where the current state-of-the-art methods require paired shadow and shadow-free images as the training data.
%
This paper presents a new method for shadow removal using unpaired data, enabling us to avoid tedious annotations and obtain more diverse training samples.
%by learning the intrinsic statistics of shadow and shadow-free images from unpaired training data.
%
However, directly employing adversarial learning and cycle-consistency constraints is insufficient to learn the underlying relationship between the shadow and shadow-free domains, since the mapping between shadow and shadow-free images is not simply one-to-one.
%the mapping function between shadow and shadow-free images is not simply deterministic.
%
%
%
%o adopt the adversarial learning method and cycle-consistency constraints to learn the intrinsic statistics and underlying relationship between the shadow and shadow-free domains. 
%
%However, we can have infinite forms of shadows on the shadow-free images, which will destroy the assumption of cycle-consistency constrain that considers the mapping function between two domains is approximately deterministic.
%
To address the problem, we formulate Mask-ShadowGAN, a new deep framework that automatically learns to produce a shadow mask from the input shadow image and then takes the mask to guide the shadow generation via re-formulated cycle-consistency constraints.
%By this means, we re-formulate the cycle-consistency constraints to optimize the framework.
Particularly, the framework simultaneously learns to produce shadow masks and learns to remove shadows, to maximize the overall performance.
%In the training process, the quality of shadow masks increases with the quality of shadow removal results.
% to add and remove shadows from shadow-free and shadow images.
%\phil{some technical things... pull in one/two sentences from sec 3}
%
Also, we prepared an unpaired dataset for shadow removal and demonstrated the effectiveness of Mask-ShadowGAN on various experiments, even it was trained on unpaired data.
%new shadow removal dataset including unpaired shadow and shadow-free images and evaluate our method on this dataset as well as two common shadow removal datasets.
%
%Experimental results demonstrate the effectiveness of Mask-ShadowGAN over others.
%
\end{abstract}

\input{section1-introduction}
\input{section2-relatedwork}
\input{section3-methods}
\input{section4-dataset}
\input{section5-experiments}
\input{section6-conclusion}

%\section*{Acknowledgments}
\vspace*{-3mm}
\paragraph{Acknowledgments.}
This work was supported by the National Basic Program of China, 973 Program (Project no. 2015CB351706), the Shenzhen Science and Technology Program (JCYJ20170413162256793 \& JCYJ20170413162617606), the Research Grants Council of the Hong Kong Special Administrative Region (Project no. CUHK 14201918), and the CUHK Research Committee Funding (Direct Grants) under  project code - 4055103.
%
%the National Basic Program of China, 973 Program (Project no. 2015CB351706), the Shenzhen Science and Technology Program (Project no. JCYJ20170413162617606), the Hong Kong Research Grants Council (Project no. CUHK 14225616 \& CUHK 14203416), and the CUHK Direct Grant for Research 2018/2019.
Xiaowei Hu is funded by the Hong Kong Ph.D. Fellowship.

{\small
\bibliographystyle{ieee_fullname}
\bibliography{egbib}
}

\end{document}

%% file: section1-introduction.tex
\section{Introduction}
\label{sec::introduction}

%%%%%%%%%%%%%%%%%%%%%%%%%%%%%%%%%%%%%%%%%%%%%%%
% Background
%%%%%%%%%%%%%%%%%%%%%%%%%%%%%%%%%%%%%%%%%%%%%%%
%Shadows are very common in natural images, which, however, degrade the performance of computer vision algorithms and impair the visual quality of photographs~\cite{khan2016automatic}. 
%
%Therefore, shadow removal is extraordinary useful in both computer vision and computer graphics.

Shadow removal is a very challenging task.
We have to remove the shadows, and simultaneously, restore the background behind the shadows.
Particularly, shadows have a wide variety of shapes over a wide variety of backgrounds.
%
%Recently, the statistical learning-based methods~\cite{gryka2015learning,guo2013paired,vicente2018leave}, especially the deep learning-based methods~\cite{hu2018direction,khan2016automatic,qu2017deshadownet,wang2018stacked} have achieved state-of-the-art performance on the benchmark datasets~\cite{qu2017deshadownet,wang2018stacked}. 
%for shadow removal by learning the mapping functions from shadow images to shadow-free images. 
Recently, learning-based methods~\cite{gryka2015learning,guo2013paired,khan2016automatic,vicente2018leave}, especially those using deep learning~\cite{hu2019direction,qu2017deshadownet,wang2018stacked}, have become the de facto standard for shadow removal, given their remarkable performance.
These methods are typically trained on pairs of shadow and shadow-free images in a supervised manner, where
the paired data is prepared by taking a photo with shadows and then taking another photo of the scene without shadows by removing the associated objects. 

Such approach to prepare training data has several limitations.
First, \emph{it is very tedious to prepare the training data}, since for each scene, we need to manually fix the camera and then add \& remove objects to obtain a pair of shadow and shadow-free images.
Moreover, the approach \emph{limits the kinds of scenes that data can be prepared}, since it is hard to capture shadow-free images for shadows casted by large objects such as trees and buildings.
Lastly, \emph{the training pairs may have inconsistent colors and luminosity, or shift in camera views}, since the camera exposure and pose, as well as the environmental lighting, may vary when we take the photo pair with and without the shadows.
%
%, which we cannot simply remove. 
%Moreover, the objects, which produce shadows, cannot appear in the shadow and shadow-free image pairs under the same light condition.\phil{I can't follow this sentence... what do you want to say? Is it the second bullet point? Or continued from the 1st point?}
%\xwhu{We have an object, which can produce shadows. This object is in both shadow and shadow-free images. This situation is impossible. But our method may not solve this problem well. Mayber it is better to ignore this situation?} 

\input{figs/teaser.tex}

%%%%%%%%%%%%%%%%%%%%%%%%%%%%%%%%%%%%%%%%%%%%%%%
% Existing work and compare
%%%%%%%%%%%%%%%%%%%%%%%%%%%%%%%%%%%%%%%%%%%%%%%
To address these problems, we present a new approach to learn to remove shadows from \emph{unpaired training data}.
%Hence, we can leverage \phil{a large set of?} training images with richer variety of shadows and scenes.
%
Our key idea is to learn the underlying relationship between a shadow domain $\mathbb{D}_s$ (a set of real images with shadows) and a shadow-free domain $\mathbb{D}_f$ (a set of real shadow-free images), where we do not have any explicit association between individual images in $\mathbb{D}_s$ and $\mathbb{D}_f$.
%instead of directly learning the mapping function from the shadow image to the corresponding shadow-free image in a pixel-to-pixel manner, we 
%
%In this scenario, 
%instead of directly learning the mapping function from the shadow image to the corresponding shadow-free image in a pixel-to-pixel manner, we explore the underlying relationship between the shadow domain $S$ and shadow-free domain $S_f$ at the level of a set of images.  
%
Here, we want to train a network $G_f$, which takes a shadow image as input and produces an output image that is indistinguishable from the shadow-free images in $\mathbb{D}_f$, by adversarial learning~\cite{goodfellow2014generative,isola2017image}.
This mapping is highly under-constrained, so the network can easily collapse during the training~\cite{zhu2017unpaired}.
Hence, we train another network $G_s$ to learn the inverse mapping,~\ie, $G_s$, to translate a shadow-free image into a shadow image like those in $\mathbb{D}_s$, and impose the following cycle-consistency constraints~\cite{zhu2017unpaired} on images $I_s \in \mathbb{D}_s$ and $I_f \in \mathbb{D}_f$,~\ie, $G_s(G_f(I_s))$ should be the same as the input shadow image $I_s$, and
$G_f(G_s(I_f))$ should be the same as the input shadow-free image $I_f$.

%%%%%%%%%%%%%%%%%%%%%%%%%%%%%%%%%%%%%%%%%%%%%%%
% What is the underlying observation behind our method
%%%%%%%%%%%%%%%%%%%%%%%%%%%%%%%%%%%%%%%%%%%%%%%

Fundamentally, deep neural networks generate a unique output for the same input.
Having said that, for the same shadow-free image, $G_s$ always generates the same shadow image with the same shadow shape.
However, this is clearly insufficient to shadow generation, since for the same background, we may have different shadows.
Figure~\ref{fig:teaser}(a) shows a further illustration: starting from different shadow images with the same background, while $G_f$ produces the same shadow-free image, $G_s$ will fail to generate different outputs that match the corresponding original inputs.
Hence, the cycle-consistency constraint cannot hold, and we cannot train $G_f$ and $G_s$ to learn to remove and generate shadows.

%Unfortunately, \emph{the relation between the shadow-free image and shadow image is not a one-to-one mapping}. 
%
%On a shadow-free image, there are infinite forms of shadows that can be added to the image.
%%Given a shadow-free image, we can add any numbers of shadow regions with any shapes on any image positions. 
%%
%However, the cycle-consistency constraint implicitly assumes that the network associates each input image with a unique output %image.
%Hence, 
%(i) given a real shadow-free image, the network $G_s$ will produce a unique shadow image with a particular style of shadow %regions.
%%
%(ii) given multiple real shadow images, where different forms of shadows are cast on the same background, the network $G_f$ %should generate the same shadow-free image. 
%Taking the generated shadow-free image as input, the network $G_s$ will fail to produce multiple shadow images to match the %contents of the real shadow images; see the schematic illustration in Figure~\ref{fig:teaser}(a).
%%, these shadow images have the same shadow-free image. 
%%
%%
%In this situation, the cycle-consistency constraint does not hold and the network optimization will fail to make progress.

To address the problem, we formulate a mask-guided generative adversarial network, namely \emph{Mask-ShadowGAN}, which learns to produce a shadow mask from an input shadow image during the training and takes the mask to guide $G_s$ to generate shadows.
%
%To take the diversity of shadows into account, in this paper, we present a framework, named as \emph{Mask-ShadowGAN}, which first learns to provide the shadow masks from a set of shadow images, and then uses the shadow masks as the guidance to indicate how to generate shadows on the shadow-free images.
% to generate shadow images, where the appearances of shadows are corresponding to the shadow masks. 
%
Therefore, given a shadow-free image, we can generate different shadows by using different shadow masks.
Further, from an input shadow image, we can produce a shadow-free image and generate suitable shadows on the image to produce an output that matches the corresponding input; see Figure~\ref{fig:teaser}(b).
Hence, we can adopt the cycle-consistency constraint to train $G_f$ and $G_s$.

To our best knowledge, this is the first data-driven method, which trains a network with unpaired data, for shadow removal.
Particularly, we design Mask-ShadowGAN to learn to remove shadows from unpaired data, bringing forth the advantage of using more shadow and shadow-free images for network learning. 
%which brings shadow removal closer to practical usage.
%
Also, we prepare the first unpaired shadow-and-shadow-free image dataset with diverse scenes.
% shadow images.
%with more practical scenes to evaluate the performance of shadow removal algorithms. 
%
Last, we perform various experiments to evaluate Mask-ShadowGAN and demonstrate its effectiveness, even it was trained on unpaired data.
%on the current shadow removal datasets and also our dataset with unpaired images.
%
Results show that our method produces comparable performance with existing works on the existing benchmarks, and outperforms others for more general shadow images without paired ground truth images. 
The source code and dataset are publicly available at {\small \url{https://xw-hu.github.io/}}.

%
%Below, we summarize the major contributions of this work:
%\begin{itemize}[]
%	%
%	\item
%	First, to our best knowledge, this work is the first data-driven method for shadow removal trained with unpaired image %data.
%	%
%	\item
%	Second, we present a mask-guided generative adversarial network (Mask-ShadowGAN) for learning to remove shadows from %unpaired data, which brings the shadow removal closer to practical applications.
%	%Second, we present a novel framework for shadow removal by first learning the shadow masks from shadow images and then %using the shadow masks to guided the generated or real shadow-free images in the adversarial learning. 
%	%
%	%This framework makes learning from unpaired shadow/shadow-free data into practice and brings the shadow removal closer to %practical applications.
%	%
%	\item
%	Third, we construct the first unpaired shadow/shadow-free dataset, which provides the shadow images with more practical %scenes to evaluate the performance of shadow removal algorithms. 
%	%
%	\item
%	Last, we perform various experiments to evaluate our Mask-ShadowGAN on the current shadow removal datasets and also our %dataset with unpaired images.
%	%
%	Results show that our method can achieve comparable performance with the existing works trained with paired images on the %current datasets, and outperforms these works on the shadow images without paired ground truth images. 
%	
%\end{itemize}

%% file: figs/teaser.tex
%%%%%%%%%%%%%%%%%%%%%%%%%%%%%%%%%%%%%%%%%%%%%%%%%%%%%%%%%%%%%%%%%%%%
% teaser
%%%%%%%%%%%%%%%%%%%%%%%%%%%%%%%%%%%%%%%%%%%%%%%%%%%%%%%%%%%%%%%%%%%%

\begin{figure}[t!]
\centering
\vspace{-1.5mm}
\includegraphics[width=0.89\linewidth]{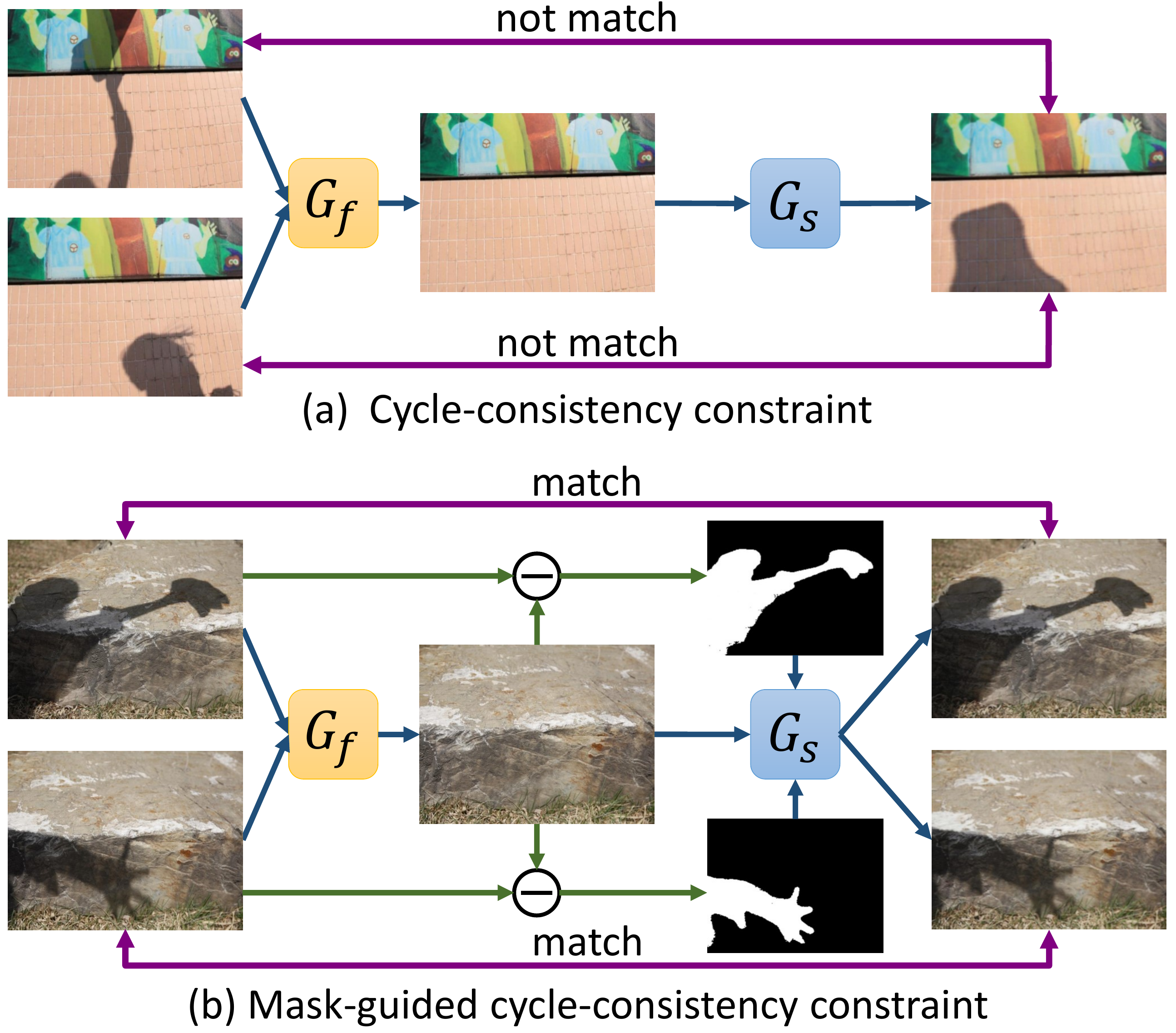}
\vspace{-1.5mm}
%\vspace{-2.5mm}
\caption{Directly using (a) the cycle-consistency constraint is insufficient; the generator $G_s$ cannot produce different shadow image outputs for different inputs.
(b) Mask-ShadowGAN learns a shadow mask from the input to guide $G_s$ to generate shadow images that better match the inputs.}
%
%\caption{In these examples, (a) cycle-consistency constraint fails, since the generator $G_s$ is unable to produce different images to match multiple input images; (b) under the guidance of shadow masks, the  generator $G_s$ can produce multiple images with various forms of shadows to match the input images.}
\label{fig:teaser}
%\vspace{-4mm}
\vspace{-3.5mm}
\end{figure}
%\phil{It looks to me that the teaser can be slightly smaller like this}

%% file: section2-relatedwork.tex
\section{Related Work}
\label{sec:related_work}

\subsection{Shadow Removal}
Early methods remove shadows by modeling images as combinations of shadow and shadow-free layers~\cite{arbel2011shadow,finlayson2009entropy,finlayson2002removing,finlayson2006removal,fredembach2005hamiltonian,liu2008texture,mohan2007editing,yang2012shadow}, or by transferring colors from non-shadow to shadow regions~\cite{shor2008shadow,wu2005bayesian,wu2007natural,xiao2013fast}. 
Since the underlying shadow models are not physically correct, they usually cannot handle shadows in complex real scenes~\cite{khan2016automatic}.
Later, statistical-learning methods were explored to find and remove shadows using features, such as intensity~\cite{gong2014interactive,gryka2015learning,guo2013paired}, color~\cite{guo2013paired,vicente2018leave}, texture~\cite{guo2013paired,vicente2018leave}, and gradient~\cite{gryka2015learning}.
However, these hand-crafted features lack high-level semantics for understanding the shadows. 
Later, Khan~\etal~\cite{khan2014automatic,khan2016automatic} adopted convolutional neural networks (CNNs) to detect shadows followed by a Bayesian model to remove shadows. 

In recent years, shadow removal mainly relies on CNNs that are trained end-to-end to learn the mapping from paired shadow and shadow-free images.
% in large training data.
%
Qu~\etal~\cite{qu2017deshadownet} developed three subnetworks to extract features from multiple views and embedded these subnetworks in a framework to remove shadows.
Wang~\etal~\cite{wang2018stacked} used a conditional generative adversarial network (CGAN) to detect shadows and another CGAN to remove shadows.
Hu~\etal~\cite{hu2019direction,Hu_2018_CVPR} explored the direction-aware spatial context to detect and remove shadows.
%
%\phil{the most recent ones, including those on detection... one sentence}
%
However, these methods are all trained on paired images, which incur several limitations, as discussed in the introduction.
Similarly, the recent shadow detection methods~\cite{vicente2016large,le2018a+d,zhu2018bidirectional,zheng2019distraction} also trained their deep neural networks using paired data.
%limits the available scenes in the real world. 
%
%Moreover, the training pairs may have inconsistent colors, illumination~\cite{hu2018direction} and geometry, due to the variance of camera exposure and pose as well as environmental lighting in the data collection process.
%
Unlike previous works, we present a new framework based on adversarial learning to learn to remove shadows from unpaired training data.

%%%%%%%%%%%%%%%%%%%%%%%%%%%%%%%
\input{figs/arc.tex}
%%%%%%%%%%%%%%%%%%%%%%%%%%%%%%%

\subsection{Unsupervised Learning}
Unsupervised learning receives great attention in recent years.
These methods can roughly be divided into three approaches.
The first approach learns the feature representations by generating images, where early methods, such as the auto-encoder~\cite{hinton2006reducing} and denoising auto-encoder~\cite{vincent2008extracting}, encoded the input images in a latent space by reconstructing them with a low error.
Some more recent works transferred the synthetic images into ``real'' images through adversarial learning~\cite{shrivastava2017learning}.
The second approach is self-supervised learning, which learns the invariant features by designing the auxiliary training objectives via labels that are free to obtain,~\eg, Doersch~\etal~\cite{doersch2015unsupervised} predicted the location of image patches for feature learning;
Pathak~\etal~\cite{pathak2017learning} learned the feature representations from segmented moving objects in videos; and
Wang~\etal~\cite{wang2017transitive} learned the visual representations from the transitive relations between images.

The last approach is more related to our work.
It learns the underlying mapping between domains in the form of unpaired data. 
CycleGAN~\cite{zhu2017unpaired} and other similar models~\cite{kim2017learning,yi2017dualgan} used two generative adversarial networks~\cite{goodfellow2014generative} to formulate the cycle consistency constraints, and learned the mapping to translate images between domains.
Though they can learn an arbitrary one-to-one mapping, a trained network can only produce a single output for the same input image.
To extend cycle consistency to handle many-to-many mapping, Almahairi~\etal~\cite{DBLP:conf/icml/AlmahairiRSBC18} and Lee~\etal~\cite{lee2018diverse} explored the use of latent variables to generate diverse outputs.
In contrast, the relationship between shadow and shadow-free images can be explicitly modelled by their difference.
%, shadow removal is a many-to-one mapping problem, while shadow generation is one-to-many.
%, since there are infinite ways of translating a shadow-free image to shadow image.
Hence, in this work, we design a framework to learn to produce the shadow mask for guiding the shadow generation, and learn to use the difference between shadow and shadow-free images in the shadow regions to model the relationship between the shadow and shadow-free images.

%In contrast, shadow removal is many-to-one (or one-to-many) mapping problem, where we can add infinite forms of shadows to the shadow-free images and the shadow image has its unique shadow-free image. 
%The shadow regions are the difference between shadow image and shadow-free image, and we can directly use the difference as the guidance to model the relationship between shadow and shadow-free images.

%% file: figs/arc.tex
%%%%%%%%%%%%%%%%%%%%%%%%%%%%%%%%%%%%%%%%%%%%%%%%%%%%%%%%%%%%%%%%%%%%
% arc
%%%%%%%%%%%%%%%%%%%%%%%%%%%%%%%%%%%%%%%%%%%%%%%%%%%%%%%%%%%%%%%%%%%%

\begin{figure*} [t!]
	\centering

	\includegraphics[width=0.95\textwidth]{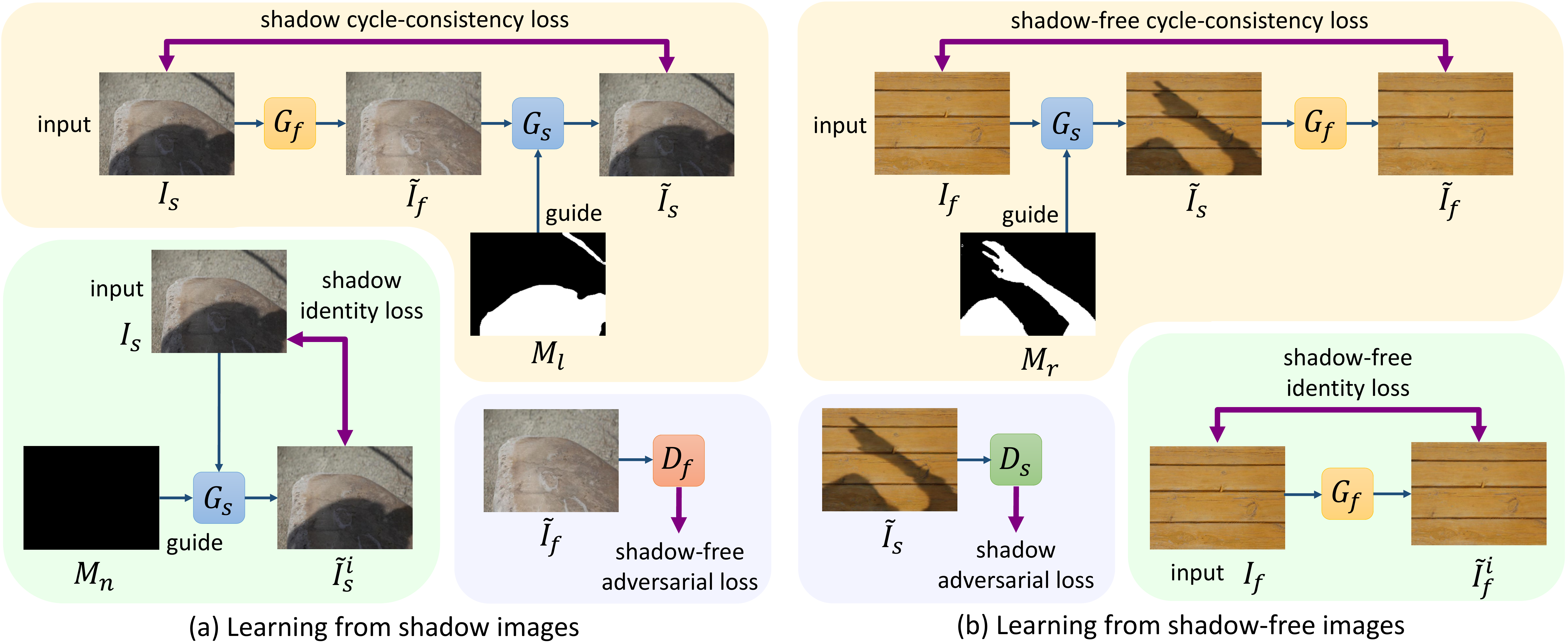} 
	\vspace{-2mm}
	\caption{The schematic illustration of our Mask-ShadowGAN, which has two parts: (a) one to
		learn from real shadow images and (b) the other to learn from real shadow-free images. 
		Each part includes three losses: cycle-consistency loss (yellow), identity loss (green), and adversarial loss (blue).
	    Besides, $G_f$ and $G_s$ denote the generators, which produce the shadow-free and shadow images while $D_f$ and $D_s$ are the discriminators to determine the whether the generated images are real shadow-free or shadow images. $I_s$ and $I_f$ are the real shadow and shadow-free image; $\tilde{I}_s$ and $\tilde{I}_s^i$ denote the generated shadow images while $\tilde{I}_f$ and $\tilde{I}_f^i$ denote the generated shadow-free images; $M_n$, $M_l$ and $M_r$ are the shadow masks.}
	\label{fig:arc}
	\vspace{-3mm}%\phil{don't reduce too much}
\end{figure*}

%% file: section3-methods.tex
%%%%%%%%%%%%%%%%%%%%%%%%%%%%%%%%%%%%%%%%%%%%%%%%%%%%%%%%%%%%%%%%%%%%
% Method
%%%%%%%%%%%%%%%%%%%%%%%%%%%%%%%%%%%%%%%%%%%%%%%%%%%%%%%%%%%%%%%%%%%%

\section{Methodology}
\label{sec:method}

%Our goal is to learn the underlying relationship between shadow domain $\mathbb{D}_S$ and shadow-free domain $\mathbb{D}_F$, and %then apply the learned relationship to remove shadows.
%%
%%To achieve this, we need to learn two generative networks (generators): one transforms the shadow images to shadow-free images %while another transforms the shadow-free images to shadow images.
%%
%Figure~\ref{fig:arc} outlines the framework of our MaskShadowGAN, which has two parts:
%one part to learn to optimize the networks from real shadow images, and 
%another to learn from real shadow-free images. 
%%
%We will elaborate these two parts in the following subsections.

%To learn to remove shadows based on the underlying relationship between shadow domain $\mathbb{D}_s$ and shadow-free domain $\mathbb{D}_f$, we develop the Mask-ShadowGAN framework outlined in Figure~\ref{fig:arc}.
%Our framework has two parts:
%one part to learn from real shadow images (Section~\ref{sec:part1}), and 
%another to learn from real shadow-free images (Section~\ref{sec:part2}).

Figure~\ref{fig:arc} outlines the overall network architecture of our Mask-ShadowGAN framework, which has two parts:
one to learn from real shadow images (Section~\ref{sec:part1}) and the other to learn from real shadow-free images (Section~\ref{sec:part2}).

%%%%%%%%%%%%%%%%%%%%%%%%%%%%%%%%%%%%%%%%%%%%%%%%%%%%%%%%%%%%%%%%%%

\subsection{Learning from Shadow Images}
\label{sec:part1}
%As shown in Figure~\ref{fig:arc}(a), first, we start from the real shadow image $S$ and use one generator $G_f$ to transform this image to the shadow-free image $\tilde{S}_f$. 
%Then, we use an adversarial discriminator $D_f$ to distinguish whether it is a real shadow-free image or not:
%
Starting from a real shadow image $I_s$, we first use a generator network $G_f$ to transform it into a shadow-free image $\tilde{I}_f$.
Then, we use an adversarial discriminator $D_f$ to differentiate whether $\tilde{I}_f$ is a real shadow-free image or not:
\begin{equation}
\tilde{I}_f = G_f(I_s), \ D_f(\tilde{I}_f) = \text{real or fake ?} 
\end{equation}
Here, we optimize the following objective function, simultaneously for the generator and its discriminator:
\begin{equation}
\begin{aligned}
L_{GAN}^a(G_f, D_f) \ = \ \mathcal{E}_{I_f \sim p_{data}(I_f)}[\log(D_f(I_f))] 
\\
\ + \ \mathcal{E}_{I_s \sim p_{data}(I_s)}[\log(1 - D_f(G_f(I_s))] \ ,
\end{aligned}
\end{equation}
%\phil{use f instead of a for the superscript?}
%\xwhu{but here is in the subsection "learning from shadow image". "s" or "f" are both confused.}
%\phil{can't follow the details in the equation... please fix the symbols carefully and define them clearly, e.g., what is $p_{data}$?}
%
%\if 0
%\begin{eqnarray}
%L_{GAN}^a(G_{S_f}, D_{S_f}) 
%& = &
%\mathbb{E}_{S_f \sim p_{data}(S_f)}[\log(D_{S_f}(S_f))]
%\nonumber
%\\
%& + &
%\mathbb{E}_{S \sim p_{data}(S)}[\log(1 - D_{S_f}(G_{S_f}(S))] \ ,
%\end{eqnarray}
%\fi
%
where $\mathcal{E}$ denotes the error;
$p_{data}$ denotes the data distribution; and
%$I_f \sim p_{data}(I_f)$ means $I_f$ should be selected from the distribution of the shadow-free images in $\mathbb{D}_f$.
%\phil{why not write $I_f \sim p_{data}(\mathbb{D}_f)$? It looks easier to follow, since we have defined $\mathbb{D}_f$}
%\xwhu{$\mathbb{D}_f$ is similar as D_f, and other GAN paper uses the same formula, like S \sim p_{data}(S). }
%
%\phil{is $I_f \sim p_{data}(I_f)$ correct? shall it be $\tilde{I}_f \sim p_{data}(I_f)$?}
%\xwhu{$I_f \sim p_{data}(I_f)$ is correct. It means we select the real image from this distribution (dataset).}
%\phil{please check and revise sec 3 carefully... I will try to continue 45 min. later}
$I_f \sim p_{data}(I_f)$ and $I_s \sim p_{data}(I_s)$ indicate that $I_f$ and $I_s$ are selected, respectively from the data distribution $p_{data}$ over the shadow-free and shadow datasets.
However, if we use the adversarial loss alone to optimize the generator, there may exist some artifacts on the generated images~\cite{isola2017image}, which may also successfully fool the discriminator~\cite{zhu2017unpaired}. 
Hence, we take another generator $G_s$ to transform the generated shadow-free image back to its original shadow image and encourage their contents to be the same. 
%\phil{BTW, I tried to simplify the symbol notations to make it easier to read... please try to follow, and update Fig.1 \& 2 as well... let me work on the paper tomorrow?}

As presented earlier in the introduction, we can produce multiple shadow images from one shadow-free image by adding the shadow regions of different shapes at different image locations.
To preserve the consistency between the generated shadow image and the original one, we use a shadow mask $M_l$ as the guidance to indicate the shadow regions, and concatenate the shadow mask $M_l$ with the generated shadow-free image $\tilde{I}_f$ as the input to the generator $G_s$, which produces the shadow image $\tilde{I}_s$:
\begin{equation}
\tilde{I}_s = G_s(\tilde{I}_f, M_l) \ ,
\end{equation}
where the shadow mask $M_l$ is the difference between the real shadow image $I_s$ and the generated shadow-free image $\tilde{I}_f$.
The shadow mask is a binary map, where zeros indicate non-shadow regions and ones indicate shadow regions; see Section~\ref{mask_generation} for the details.
Then, we formulate the following shadow cycle consistency loss to encourage the reconstructed image $\tilde{I}_s$ to be similar to the original input real shadow image $I_s$, and optimize the mapping functions in $G_s$ and $G_f$ by a cycle-consistency constraint:
\begin{equation}
L_{cycle}^a(G_f, G_s) = \mathcal{E}_{I_s \sim p_{data}(I_s)}[||G_s( G_f(I_s), M_l ) - I_s||_1] \ . 
\end{equation}
%\begin{align}
%&L_{cycle}^a(G_{S_f}, G_{S}) = 
%\nonumber
%\\
%&\mathcal{E}_{S \sim p_{data}(S)}[||G_{S}( G_{S_f}(S) ) - S||_1]. 
%\end{align}
%
By leveraging the $L_1$ loss $||.||_1$ to calculate the difference on each pixel, generator $G_s$ will learn to produce a shadow image as well as to capture the relationship between the shadow image and shadow mask,~\ie, zeros denote the non-shadow regions, while ones denote the shadow regions.
%; see $\tilde{I}_s$ and $M_l$ in Figure~\ref{fig:arc}(a).
%will learn to produce a shadow mask in the form of a binary map; see $M_l$ in Figure~\ref{fig:arc}(a).
%the generator $G_s$ can learn to know which value ($0 \ \text{or} \ 1$) on the shadow mask represents the non-shadow or shadow regions.
 
See again the Figure~\ref{fig:arc}(a), 
we further use a mask $M_n$ with all zero values and the real shadow image $I_s$ as the input of $G_s$, and generate an image $\tilde{I}_s^i$, which contains no newly added shadows: 
\begin{equation}
\tilde{I}_s^i = G_s(I_s, M_n) \ .
\end{equation}
Then, we leverage the shadow identity loss~\cite{taigman2016unsupervised} to regularize the
output to be close to the input shadow image:
\begin{equation}
L_{identity}^a(G_s) = \mathcal{E}_{I_s \sim p_{data}(I_s)}[||G_s( I_s, M_n ) - I_s||_1] \ . 
\end{equation}
Hence, we can encourage that no shadows will be added on the input shadow image in the generated image $\tilde{I}_s^i$ under the guidance of $M_n$, and we can also preserve the color composition between the input and output images~\cite{zhu2017unpaired}.

\subsection{Learning from Shadow-free Images}
\label{sec:part2}
Figure~\ref{fig:arc}(b) shows the framework on how to learn from the shadow-free images for shadow removal. 
Given a real shadow-free image $I_f$, we use a generator $G_s$ to produce the shadow image $\tilde{I}_s$, which is used to fool the discriminator $D_s$, and makes it hard to distinguish whether it is a real shadow image or not.
As mentioned before, for the generator $G_s$, we need a shadow mask as the input to indicate the shadow regions. 
Here, we are able to use a mask $M_r$ with any forms of shadows as the guidance and produce the generated shadow image $\tilde{I}_s$:
%
%There are infinite forms of shadows that we can add to the shadow-free image, and we randomly select on mask $M_r$, which is learned from the real shadow images; see Section~\ref{sec:part1}. 
%
%Since the shadow mask is 
%Unlike the fixed types of shadow masks used in the above subsection, here, we can use the shadow mask, where the shadows have any shapes and locations.
%\phil{this sentence is hard to follow}
%
%To make the generated shadow regions look real, we adopt the shadow masks learned from the real shadow images (Section~\ref{sec:part1}), and then randomly select one mask $M_r$ as the input for the generator $G_s$:
%
\begin{equation}
\tilde{I}_s = G_s(I_f, M_r), \ D_s(\tilde{I}_s) = \text{real or fake ?} 
\end{equation}
To make the generated shadow regions look real, we randomly select one shadow mask learned from the real shadow image; see Section~\ref{mask_generation} for the details. 
By leveraging different shadow masks as the guidance, we produce multiple shadow images with different forms of shadows.
Therefore, \emph{a large number of shadow images will be created, thus increasing the generalization capability of the deep models.}
%
%Hence, for one shadow-free image, we can produce multiple shadow images with different kinds of shadow regions.
%
Finally, we use the adversarial loss to optimize the generator $G_s$ and discriminator $D_s$:
\begin{equation}
\begin{aligned}
L_{GAN}^b(G_s, D_s) \ = \ \mathcal{E}_{I_s \sim p_{data}(I_s)}[\log(D_s(I_s))] 
\\
\ + \ \mathcal{E}_{I_f \sim p_{data}(I_f)}[\log(1 - D_s(G_s(I_f, M_r))] \ .
\end{aligned}
\end{equation}

To leverage the cycle-consistency constraint, we adopt the generator $G_f$ to produce the shadow-free image $\tilde{I}_f$ from the generated shadow image $\tilde{I}_s$, s.t., $\tilde{I}_f = G_f(\tilde{I}_s)$, and use the shadow-free cycle consistency loss to optimize the networks:
\begin{equation}
%\begin{aligned}
L_{cycle}^b(G_s, G_f) = \mathcal{E}_{I_f \sim p_{data}(I_f)}
%\\
[||G_f( G_s(I_f, M_r)) - I_f||_1] \ . 
%\end{aligned}
\end{equation}

Last, we adopt the generator $G_f$ to produce a shadow-free image $\tilde{I}_f^i$ by taking the real shadow-free image $I_f$ as the input, s.t., $\tilde{I}_f^i = G_f(I_f)$, and then use the shadow-free identity loss to force the content of the input and output images to be the same:
\begin{equation}
L_{identity}^b(G_f) = \mathcal{E}_{I_f \sim p_{data}(I_f)}[||G_f(I_f) - I_f||_1] \ . 
\end{equation}
By using the identity loss as a constraint, the generator $G_f$ will learn to remove shadows without changing colors on non-shadow regions.

\subsection{Loss Function}
In summary, the final loss function for our Mask-ShadowGAN is a weighted sum of the adversarial loss, cycle consistency loss, and identity loss in two parts of the framework:
\begin{align}
& L_{final}(G_s, G_f, D_s, D_f) \\
\nonumber
& = \omega_1 (L_{GAN}^a(G_f, D_f) + L_{GAN}^b(G_s, D_s)) \\
\nonumber
& + \omega_2 (L_{cycle}^a(G_f, G_s) + L_{cycle}^b(G_s, G_f)) \\
\nonumber
& + \omega_3 (L_{identity}^a(G_s) + L_{identity}^b(G_f)) \ .
\end{align}
We follow~\cite{zhu2017unpaired} and empirically set $\omega_1$, $\omega_2$, and $\omega_3$ as $1$, $10$, and $5$, respectively. Finally, we optimize the whole framework in a minimax manner:
\begin{equation}
%G_S^*, G_{S_f}^* = 
\arg \min_{G_s, G_f} \max_{D_s, D_f} L_{final}(G_s, G_f, D_s, D_f).
\end{equation}

\subsection{Mask Generation}
\label{mask_generation}
As described earlier, we design a shadow mask to indicate how to generate shadows on the shadow-free images. 
We obtain the shadow mask $M$ by calculating the difference between the real shadow image $I_s$ and the generated shadow-free image $\tilde{I}_f$, and then binarizing the result:
\begin{equation}
M = \mathbb{B}( \ \tilde{I}_f - I_s \ , \ t \ ) \ ,
\end{equation}
where $\mathbb{B}$ indicates the binarization operation, which sets the pixels as one, when their values are greater than the threshold $t$, otherwise, as zero.
We obtain the threshold $t$ by Otsu's algorithm~\cite{otsu1979threshold}, which calculates the optimum threshold to separate shadow and non-shadow regions by minimizing the intra-class variance.

Since we obtain one shadow mask from one real shadow image, we adopt a list to save multiple shadow masks produced from the shadow images in one dataset.
During the training process, the quality of shadow masks increases with the quality of generated shadow-free images $\tilde{I}_f$. 
Hence, we update the list of shadow masks by pushing the newly generated mask (high quality) and removing the least recently added mask (low quality). 
This process is achieved by the \emph{Queue} dataset structure, which obeys the rule of ``first in, first out''.
Moreover, we empirically set the length of the list as a quarter of image numbers in the real shadow dataset.

When accessing the shadow masks, we set $M_l$ in Figure~\ref{fig:arc}(a) as the newly generated mask from its input shadow image, and randomly choose one mask from the list as $M_r$ in Figure~\ref{fig:arc}(b).

\subsection{Network Architecture and Training Strategy}

\paragraph{Network architecture.} \
We take the network architecture designed by Johnson~\etal~\cite{johnson2016perceptual} as our generator network, which includes three convolution operations, followed by nine residual blocks with the stride-two convolutions and two deconvolutions for feature map upsampling. 
In this network, instance normalization~\cite{ulyanov2016instance} is used after each convolution and deconvolution operation.
The generator $G_f$ takes the shadow image with the channel number of three as the input, while the generator $G_s$ adopts the concatenation of the shadow-free image and shadow mask as the input, which has four channels in total. 
Both $G_f$ and $G_s$ produce a residual image with three channels, which is added with the input image as the final shadow-free or shadow image. 
For the discriminator $D_f$ and $D_s$, we use the PatchGAN~\cite{isola2017image} to distinguish whether the image patches are real or fake. 

\vspace*{-3mm}
\paragraph{Training strategy.} \
We initialized the parameters in all generators and discriminators by random noise, which follows a zero-mean Gaussian distribution with standard deviation set as $0.02$. 
Moreover, Adam~\cite{kingma2014adam} was used to optimize our networks with the first and second momentum values set as $0.5$ and $0.999$, respectively. 
We empirically set the basic learning rate as $2 \times 10^{-4}$ for the first $100$ epochs, gradually reduced it to zero with a linear decay rate in the next $100$ epochs, and then stopped the learning. 
Lastly, we built our model on PyTorch with a mini-batch size of one, and randomly cropped images for data argumentation.
%, and set the input size of $400 \times 400$, following the recent work~\cite{hu2019direction} for shadow removal.

%% file: section4-dataset.tex
\section{Unpaired Shadow Removal Dataset - USR}
\label{dataset}

\begin{figure} [tp]
	\centering
	\includegraphics[width=0.98\linewidth]{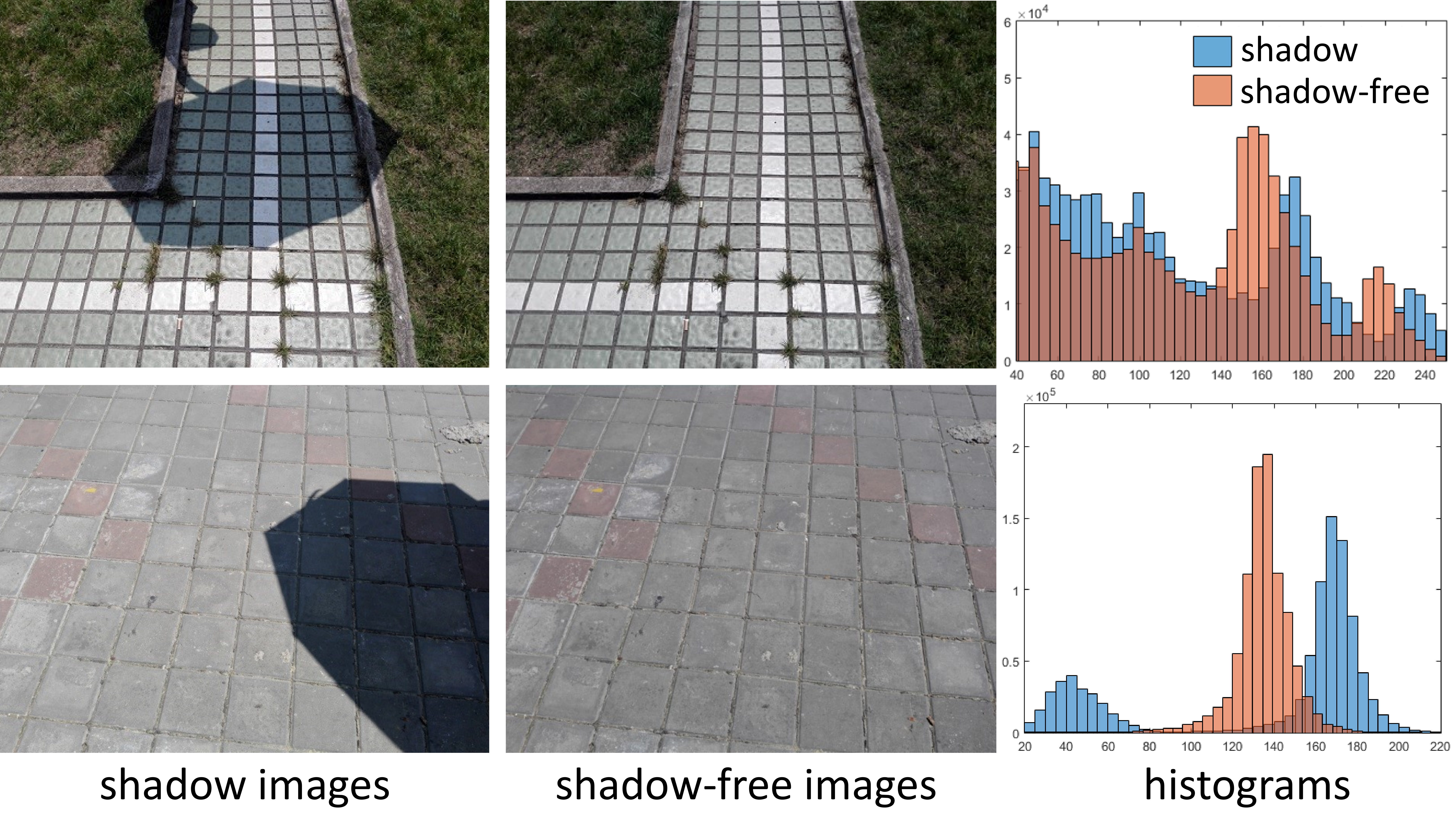}
	\vspace*{-1.5mm}
	\caption{Typical paired shadow and shadow-free images from~\cite{wang2018stacked}; note the color inconsistencies revealed by the intensity distributions in the histograms.}
	\label{fig:color_inconsistent}
	\vspace*{-3.5mm}
\end{figure}

%\phil{I tried to shorten this paragraph but this may or may not be necessary}
Existing shadow removal datasets~\cite{qu2017deshadownet,wang2018stacked} are paired.
Typically, we have to fix the camera, take a photo with shadows, then take another one without shadows by removing the associated objects.
%
%Hence, we can obtain the paired shadow/shadow-free images.
%
Due to the varied environment light and camera exposure, a training pair may have inconsistent colors and luminosity; see examples in Figure~\ref{fig:color_inconsistent}.
% with the associated histograms.
%
Also, paired data is available only for limited scenes, thus affecting the generality and practicality of the trained models.

\if 0
Existing shadow removal datasets~\cite{qu2017deshadownet,wang2018stacked} are paired.
Typically, we have to fix the camera, take a photo with shadows, and then take another photo without shadows by removing the associated objects.
%
%Hence, we can obtain the paired shadow/shadow-free images.
%
As discussed earlier, due to the varied environmental light and camera exposure, a training pair may have inconsistent colors and luminosity; see Figure~\ref{fig:color_inconsistent} for two examples.
% with the associated histograms.
%
Also, paired training data is available only for limited scenes, thus affecting the generality and practicality of the trained models.
\fi

We prepared an unpaired shadow removal dataset named USR with $2,445$ shadow images and $1,770$ shadow-free images.
The dataset contains a large variety of scenes with shadows cast by various kinds of objects,~\eg, trees, buildings, traffic signs, persons, umbrellas, railings,~\etc~
Very importantly, existing datasets typically cover only hundreds of different scenes (even with thousands of image samples), while ours cover {\em over a thousand different scenes\/}.
%
%Furthermore, some images include both objects and their shadows.
%
Furthermore, we divide the shadow images in the dataset randomly into $1,956$ images for training and $489$ for testing, and use all the $1,770$ shadow-free images for training (since they are not used in shadow removal testing).
%\emph{We will release the dataset upon the publication of this work.}

%Existing large-scale shadow removal datasets~\cite{qu2017deshadownet,wang2018stacked} were prepared by first fixing the camera, taking one photo of the scene with shadows, and then taking another photo without the shadows by removing the associated objects. 
%
%Hence, we can obtain the paired shadow/shadow-free images.
%
%However, due to the varied environmental luminosity and camera exposure, a training pair of shadow/shadow-free images may have inconsistent colors and luminosity;
%
%see Figure~\ref{fig:color_inconsistent} for examples from the paired datasets~\cite{qu2017deshadownet,wang2018stacked}, where the inconsistencies are revealed by the color histograms. 
%
%Moreover, the paired training data is only available for the limited scenes, which reduces the generality and practicality of the trained models in the real world.

%To evaluate the shadow removal methods on more scenes, we prepare a dataset, named as USR, which consists of unpaired shadow and shadow-free images.
%
%It includes images under a large variety of scenes, and the shadows are cast by trees, buildings, traffic signs, persons, umbrellas, railings, etc. 
%
%%Furthermore, some images include both objects and their shadows.
%
%Altogether, our USR dataset has $2,511$ shadow images and $1,772$ shadow-free images, where we divide the shadow images into the training set with $1,992$ images and testing set with $519$ images, and put all the shadow-free images into the training set.
%\emph{We will release the dataset upon the publication of this work.}

%% file: section5-experiments.tex
\section{Experimental Results}
\label{sec:experiments}

%%%%%%%%%%%%%%%%%%%%%%%%%%%%%%%%%%%%%%%%%%%%%%%%%%%

\subsection{Datasets and Evaluation Metrics}

\paragraph{Datasets.} Besides the USR dataset, we employed two recent shadow removal datasets (SRD~\cite{qu2017deshadownet} and ISTD~\cite{wang2018stacked}), which contain the paired shadow/shadow-free images and are used for training the existing shadow removal methods. 
%
%SRD contains $2,680$ training pairs and $408$ testing pairs of shadow and shadow-free images, which are captured under different illuminations with various shadow shapes over about a hundred scenes.
%ISTD contains triplets of shadow, shadow-free, and shadow mask images:
%$1,330$ training triplets and
%$540$ testing triplets, covering various shadow shapes for $135$ different scenes.

\vspace*{-3mm}
\paragraph{Evaluation metrics.}
We followed recent works~\cite{hu2019direction,qu2017deshadownet,wang2018stacked} to evaluate shadow removal performance by computing the root-mean-square error (RMSE) between the ground truth and predicted shadow-free images in LAB color space.
In general, a small RMSE indicates a better performance.
%Since we do not have the paired ground truth images for our USR

%%%%%%%%%%%%%%%%%%%%%%%%%%%%%%%%%%%%%%%%%%%%%%%%%%%

\input{./figs/USR_comp.tex}

\subsection{Comparison using USR}

First of all, we compare Mask-ShadowGAN with state-of-the-art shadow removal methods on the USR dataset.
The purpose here is to show that by leveraging unpaired data, we are able to train a network to learn to remove shadows of more variety shapes for a wider range of scenes.
%, which has unpaired shadow and shadow-free images.

\vspace*{-3mm}
\paragraph{USR training set.}
First, we trained our model on the USR training set and applied it to produce shadow-free images on the USR testing set.
Also, we applied several state-of-the-art methods to remove shadows on the USR testing set: DSC~\cite{hu2019direction}, Gong~\etal~\cite{gong2014interactive}, and Guo~\etal~\cite{guo2013paired}.
For DSC, we adopted its public implementation and trained its network on the SRD and ISTD datasets: ``DSC-S'' and ``DSC-I'' denote the model trained on the SRD dataset and on the ISTD dataset, respectively.
Since DSC requires paired shadow and shadow-free images, we cannot re-train it on the USR dataset, which is unpaired.
For the other methods, Gong~\etal and Guo~\etal, we downloaded and leveraged their public code with recommended parameters to produce the shadow-free image results.
Note that the code of ST-CGAN~\cite{wang2018stacked} and DeshadowNet~\cite{qu2017deshadownet} are not publicly available, we cannot evaluate them on the USR testing data.

%%%%%%%%%%%%%%%%%%%%%%%%%%%%%%%%%%%%%%%%%%%%%%%%%%%%%%%%%%%%%%%%%%%%%%%%%%%%%
\begin{table}[tp]
	\begin{center}
		\caption{User study results on the USR testing set.
			Mean ratings (from $1$ (bad) to $10$ (good)) given by the participants on the shadow removal results.}
		\vspace{-2mm}
		\label{table:user_study}
		\resizebox{0.85\linewidth}{!}{%
			\begin{tabular}{c|c}
				%
				%\cline{1-13}
				\hline
				Methods & Rating (mean \& standard dev.)\\
				\hline
				\textbf{Mask-ShadowGAN}  & $\textbf{6.30} \pm 2.97$ \\
				\hline
				DSC-I~\cite{hu2019direction, wang2018stacked} & $4.78 \pm 2.92$ \\
				DSC-S~\cite{hu2019direction, qu2017deshadownet} & $4.60 \pm 2.66$ \\
				Gong \emph{et al.}~\cite{gong2014interactive}  & $2.82 \pm 1.76$ \\
				Guo \emph{et al.}~\cite{guo2013paired} & $2.31 \pm 1.90$ \\
				
				\hline
		\end{tabular} }
	\end{center}
	\vspace*{-8mm}
\end{table}
%%%%%%%%%%%%%%%%%%%%%%%%%%%%%%%%%%%%%%%%%%%%%%%%%%%%%%%%%%%%%%%%%%%%%%%%%%%%%

%%%%%%%%%%%%%%%%%%%%%%%%%%%%%%%%%%%%%%%%%%%%%%%%%%%

\input{./figs/SRD_ISTD_comp.tex}

%\vspace*{-3mm}
%\paragraph{Quantitative and visual comparison.}
%As there are no ground-truth images in the unpaired USR dataset, we conducted a user study to evaluate the shadow removal results. 
As the unpaired USR dataset has no ground truths, we conducted a user study to evaluate the shadow removal results. 
First, we generate shadow-free images using Mask-ShadowGAN, as well as using DSC-I, DSC-S, Gong~\etal, and Guo~\etal, on the USR testing set (only shadow images).
Here, we recruited ten participants: six females and four males, aged $23$ to $30$ with mean $26.1$.
For each participant, we randomly selected $150$ shadow-free image results ($30$ per method), presented the results in random order to the participant, and asked the participant to rate the result in a scale from $1$ (bad) to $10$ (good).
Therefore, we obtained $300$ ratings ($10$ participants $\times$ $30$ images) per method.

Table~\ref{table:user_study} shows the results.
Mask-ShadowGAN received the highest ratings compared to other methods, showing its effectiveness to remove shadows for more diverse scenes, even it was trained just on unpaired data.
%outperforms the ratings on other methods by a large margin. 
%
Further, we performed a statistical analysis on the ratings by conducting t-tests between Mask-ShadowGAN and other methods.
All the t-test results show that our results are statistically significant (with $p < 0.001$) than the others, 
evidencing that the participants prefer our results more than those produced by other methods.
%; please see the supplemental material for the analysis details.
%by training our method on unpaired data with diverse scenes, we can achieve better performance on shadow removal.
%, even where no paired ground truth images are available under the similar scenes. 
%
Figure~\ref{fig:comparison_removal_USR} shows the visual comparisons, where Mask-ShadowGAN can more effectively remove the shadows and recover the background, while others may blur the images or fail to remove portions of the shadows.
Very importantly, our method was trained just on unpaired data.

%%%%%%%%%%%%%%%%%%%%%%%%%%%%%%%%%%%%%%%%%%%%%%%%%%%%%%%%%%%%%%%%%%%%%%%%%%%%%
\begin{table}[tbp]
	\begin{center}
		\caption{User study: participant ratings from $1$ (bad) to $10$ (good).
			Different trained models tested on the USR test set.}
		%		\caption{User study results on the USR testing set.
		%			Mean ratings (from $1$ (bad) to $10$ (good)) given by the participants.}
		\vspace{-2mm}
		\label{table:user_study2}
		\resizebox{0.95\linewidth}{!}{%
			\begin{tabular}{c|c|c}
				%
				%\cline{1-13}
				\hline
				Trained models & Training set &Rating (mean \& standard dev.)\\
				\hline
				\textbf{Ours-I} & ISTD & $\textbf{4.07} \pm 2.93$ \\
				DSC-I & ISTD & $2.38 \pm 2.12$ \\
				
				\hline
				\textbf{Ours-S} & SRD & $\textbf{3.38} \pm 2.42$ \\
				DSC-S & SRD & $2.93 \pm 2.39$ \\
				\hline
				%\textbf{ours-U} & USR (our dataset) & $\textbf{7.28} \pm 2.78$ \\			
				%\hline
				%
		\end{tabular} }
	\end{center}
	\vspace*{-7mm}
\end{table}
%%%%%%%%%%%%%%%%%%%%%%%%%%%%%%%%%%%%%%%%%%%%%%%%%%%%%%%%%%%%%%%%%%%%%%%%%%%%%

\vspace*{-3mm}
\paragraph{SRD \& ISTD training sets.}
Also, we trained our method separately on the training sets of SRD (``Ours-S'') and ISTD (``Ours-I'').
Then, we applied these models, as well as DSC-S and DSC-I to the USR test set, conducted another user study with four females \& six males (aged $22$ to $30$), and showed $150$ randomly-selected shadow-free image results ($30$ from each trained model) to each participant in random order.
Table~\ref{table:user_study2} shows the results.
Our method still outperforms the state-of-the-art (DSC), even it was trained on the same training set, since training in an unpaired manner improves the generalization capability of our model.
%This is because our model trained in an unpaired manner has better generalization capability.

\subsection{Comparison using SRD and ISTD}
Next, we compare our method with others on the SRD and ISTD dataset (paired data) using their ground truths.
%We compared our method against recent methods for shadow removal on the paired datasets, where the shadow and shadow-free image pairs are available for evaluation. 
%
Here, we trained our Mask-ShadowGAN on the SRD training set and tested it on the SRD testing set, then re-trained our model on the ISTD training set and tested the trained model on the ISTD testing set. 
%We trained our Mask-ShadowGAN on the training sets of SRD and ISTD respectively, and tested the trained models on the testing sets of SRD and ISTD.
%
Since Mask-ShadowGAN is designed to train on unpaired data, we randomly chose an image from the shadow image set and the other from the shadow-free image set per mini-batch during the training.

\input{./tables/table_SRDISTD_comparison.tex}

\begin{figure*} [tp]
	\centering
	\includegraphics[width=0.99\linewidth]{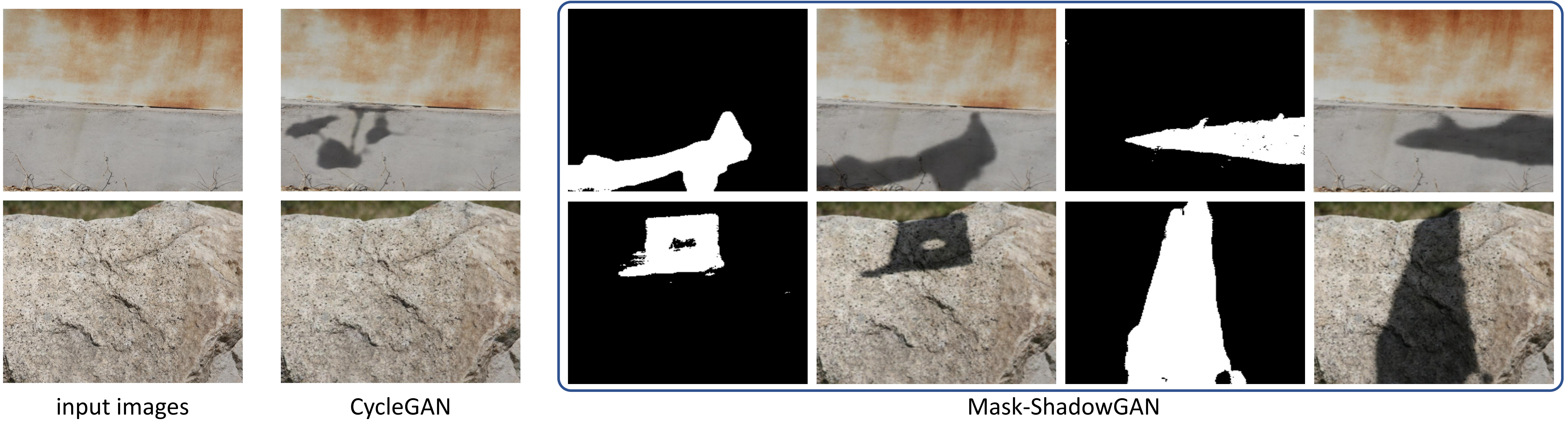}
	\vspace{-2mm}
	\caption{Comparing with CycleGAN~\cite{zhu2017unpaired} on generating shadow images.
%The binary images are the shadow masks produced in Mask-ShadowGAN.
Note that CycleGAN produces the same output for the same input, while Mask-ShadowGAN can produce different outputs, as guided by the shadow masks (binary images).}
	\label{fig:comp_cycle2}
	\vspace{-4mm}
\end{figure*}

\begin{figure} [tp]
	\centering
	\includegraphics[width=0.99\linewidth]{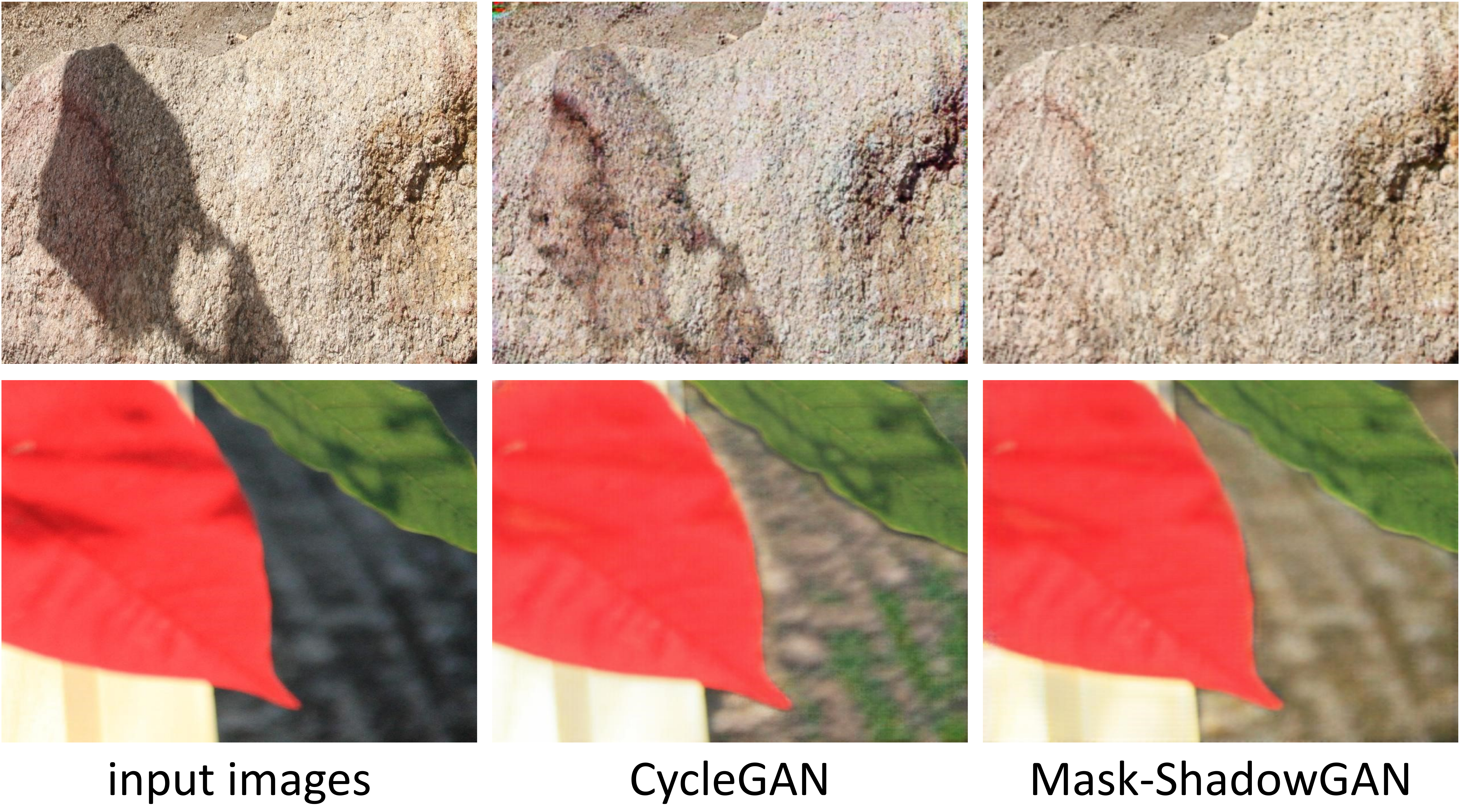}
	\vspace{-2mm}
	\caption{Visual comparison with CycleGAN~\cite{zhu2017unpaired} on generating shadow-free images.}
	\label{fig:comp_cycle1}
	\vspace{-4mm}
\end{figure}

\vspace*{-3mm}
\paragraph{Quantitative and visual comparison.}
%\vspace*{-3mm}
%\paragraph{Comparison with shadow removal methods.}
%
To compare with other shadow removal methods (DSC~\cite{hu2019direction}, ST-CGAN~\cite{wang2018stacked}, DeshadowNet~\cite{qu2017deshadownet}, Gong~\etal~\cite{gong2014interactive}, Guo~\etal~\cite{guo2013paired}, Yang~\etal~\cite{yang2012shadow}), we obtained their results directly from the authors or by generating them using the public code with the recommended parameter setting.
Here, DSC, ST-CGAN and DeshadowNet are deep networks that produce shadow-free images end-to-end, and their networks were trained on paired shadow and shadow-free images.
Gong~\etal, Guo~\etal, and Yang~\etal leverage image priors to remove shadows.

We present the results in Table~\ref{table:SRDISTD_removal}, where our method achieves RMSE values (even trained in an unpaired manner) that are comparable with those of the other deep neural networks trained on paired images, and clearly outperforms the methods based on hand-crafted features.
Note also that the code of ST-CGAN~\cite{wang2018stacked} and DeshadowNet~\cite{qu2017deshadownet} is not publicly available, and we can only report their results on the datasets used in their published papers.
%
%The methods trained on the paired shadow/shadow-free images achieve the best performance, since they 

Figure~\ref{fig:comparison_removal_SRD_ISTD} shows the visual comparison results on these two datasets, which present some challenging cases, e.g., large shadow regions (the first three rows) and shadows across the backgrounds with complex textures (the first, fourth and fifth rows). 
Although the RMSE value of Mask-ShadowGAN on these datasets is higher than the deep networks trained on paired data, Mask-ShadowGAN can generate more realistic images and better preserve the texture details occluded by shadows; see again the first, fourth and fifth rows in Figure~\ref{fig:comparison_removal_SRD_ISTD}.
This is because we learn to remove shadows from the reliable intrinsic statistics of real shadow/shadow-free images, and avoid the unrealistic grayish (blurry) outputs through adversarial learning.
%(as shown in the results of DSC, ST-CGAN and DeshadowNet).

\vspace*{-3mm}
\paragraph{Comparison with CycleGAN.}
Further, we compared our method with CycleGAN~\cite{zhu2017unpaired}, which is designed for general image-to-image translations using unpaired training data.	
Here, we adopted the author-provided implementations and used the same parameter setting as our Mask-ShadowGAN to re-train the model on the SRD and ISTD training sets.

Table~\ref{table:SRDISTD_removal} reports the results, showing that our method outperforms CycleGAN on both datasets.
By leveraging the shadow masks to guide the shadow generation for both real and generated shadow-free images, we can effectively provide constraints to the network to guide its exploration in the search space for producing more realistic results.
%It demonstrates the effectiveness of leveraging the shadow masks to guide the shadow generation for both real and generated shadow-free images, thereby further providing constraints to the network to guide its exploration in the search space and thus producing more realistic results.
%
Figure~\ref{fig:comp_cycle1} shows visual comparison results.
Mask-ShadowGAN can clearly remove the shadows, but CycleGAN tends to 
%fail to recover the regions occluded by the shadows or 
produce artifacts on the regions occluded by the shadows.
Besides, Figure~\ref{fig:comp_cycle2} shows shadow images generated from real shadow-free images by Mask-ShadowGAN and by CycleGAN.
While CycleGAN always produces the same shadow image for the same shadow-free input, Mask-ShadowGAN is able to produce multiple realistic shadow images with the help of the shadow masks, which are also learned in the network automatically from some real shadow images.

%\emph{Please see the supplemental material for more shadow removal and visual comparison results. We will release our dataset and source code, and show more results on the datasets upon the publication of this work.}

%\subsection{Evaluation on Network Design}

%% file: figs/USR_comp.tex
%%%%%%%%%%%%%%%%%%%%%%%%%%%%%%%%%%%%%%%%%%%%%%%%%%%%%%%%%%%%%%%%%%%%
% teaser
%%%%%%%%%%%%%%%%%%%%%%%%%%%%%%%%%%%%%%%%%%%%%%%%%%%%%%%%%%%%%%%%%%%%

%%%%%%%%%%%%%%%%%%%%%%%%%%%%%%%%%%%%%%%%%

\begin{figure*}[tp]
	\centering
    \vspace*{0.5mm}
    \begin{subfigure}{0.16\textwidth} %0.138\textwidth; 0.195\textwidth 0.16\textwidth
    	\includegraphics[width=\textwidth]{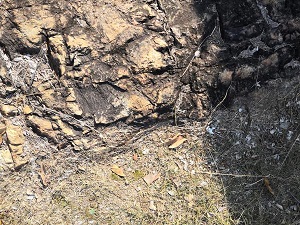}
    \end{subfigure}
    \begin{subfigure}{0.16\textwidth} %0.138\textwidth; 0.195\textwidth 0.16\textwidth
    	\includegraphics[width=\textwidth]{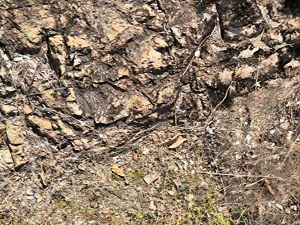}
    \end{subfigure}
    \begin{subfigure}{0.16\textwidth} %0.138\textwidth; 0.195\textwidth 0.16\textwidth
    	\includegraphics[width=\textwidth]{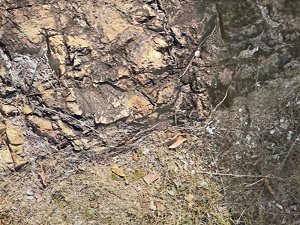}
    \end{subfigure}
    \begin{subfigure}{0.16\textwidth} %0.138\textwidth; 0.195\textwidth 0.16\textwidth
    	\includegraphics[width=\textwidth]{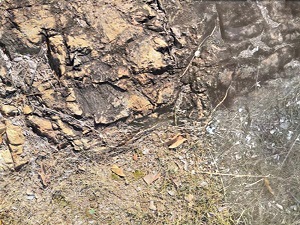}
    \end{subfigure}
    \begin{subfigure}{0.16\textwidth} %0.138\textwidth; 0.195\textwidth 0.16\textwidth
    	\includegraphics[width=\textwidth]{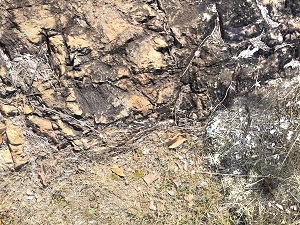}
    \end{subfigure}
    \begin{subfigure}{0.16\textwidth} %0.138\textwidth; 0.195\textwidth 0.16\textwidth
    	\includegraphics[width=\textwidth]{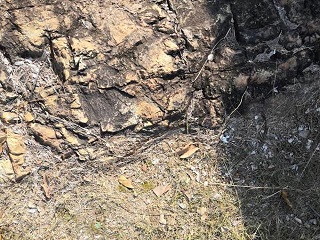}
    \end{subfigure}

    	\vspace*{0.5mm}
    \begin{subfigure}{0.16\textwidth} %0.12\textwidth; 0.195\textwidth 0.16\textwidth
    	\includegraphics[width=\textwidth]{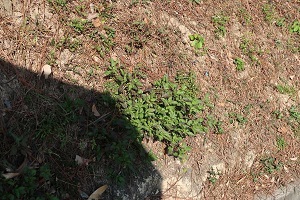}
    \end{subfigure}
    \begin{subfigure}{0.16\textwidth} %0.138\textwidth; 0.195\textwidth 0.16\textwidth
    	\includegraphics[width=\textwidth]{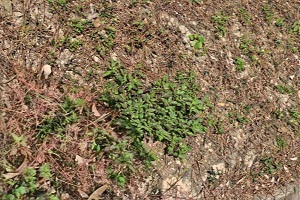}
    \end{subfigure}
    \begin{subfigure}{0.16\textwidth} %0.138\textwidth; 0.195\textwidth 0.16\textwidth
    	\includegraphics[width=\textwidth]{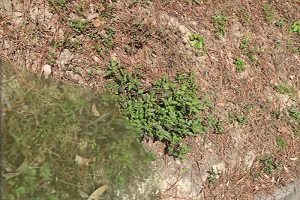}
    \end{subfigure}
    \begin{subfigure}{0.16\textwidth} %0.138\textwidth; 0.195\textwidth 0.16\textwidth
    	\includegraphics[width=\textwidth]{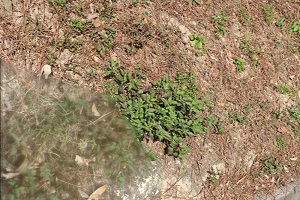}
    \end{subfigure}
    \begin{subfigure}{0.16\textwidth} %0.138\textwidth; 0.195\textwidth 0.16\textwidth
    	\includegraphics[width=\textwidth]{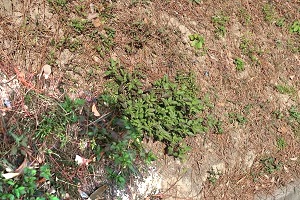}
    \end{subfigure}
    \begin{subfigure}{0.16\textwidth} %0.138\textwidth; 0.195\textwidth 0.16\textwidth
    	\includegraphics[width=\textwidth]{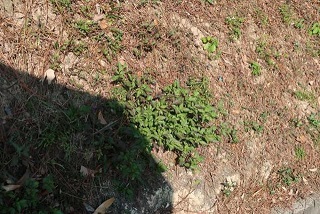}
    \end{subfigure}

    	\vspace*{0.5mm}
    \begin{subfigure}{0.16\textwidth} %0.12\textwidth; 0.195\textwidth 0.16\textwidth
    	\includegraphics[width=\textwidth]{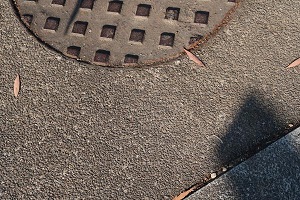}
    \end{subfigure}
    \begin{subfigure}{0.16\textwidth} %0.138\textwidth; 0.195\textwidth 0.16\textwidth
    	\includegraphics[width=\textwidth]{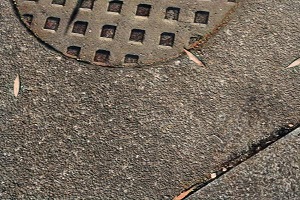}
    \end{subfigure}
    \begin{subfigure}{0.16\textwidth} %0.138\textwidth; 0.195\textwidth 0.16\textwidth
    	\includegraphics[width=\textwidth]{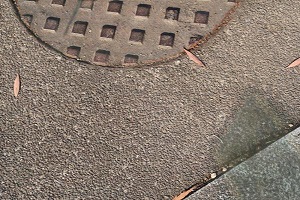}
    \end{subfigure}
    \begin{subfigure}{0.16\textwidth} %0.138\textwidth; 0.195\textwidth 0.16\textwidth
    	\includegraphics[width=\textwidth]{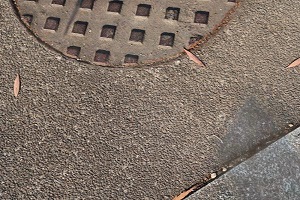}
    \end{subfigure}
    \begin{subfigure}{0.16\textwidth} %0.138\textwidth; 0.195\textwidth 0.16\textwidth
    	\includegraphics[width=\textwidth]{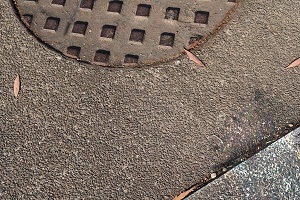}
    \end{subfigure}
    \begin{subfigure}{0.16\textwidth} %0.138\textwidth; 0.195\textwidth 0.16\textwidth
    	\includegraphics[width=\textwidth]{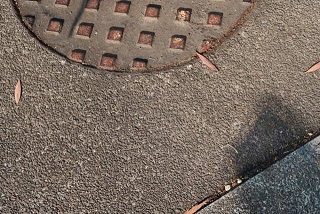}
    \end{subfigure}

	\vspace*{0.5mm}
	\begin{subfigure}{0.16\textwidth} %0.138\textwidth; 0.195\textwidth 0.16\textwidth
		\includegraphics[width=\textwidth]{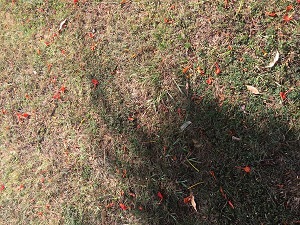}
		\vspace{-5.5mm} \caption*{{\footnotesize input}}
	\end{subfigure}
	\begin{subfigure}{0.16\textwidth}
		\includegraphics[width=\textwidth]{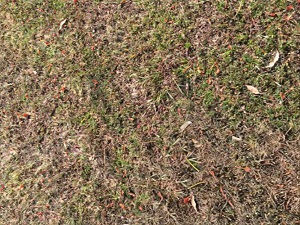}
		\vspace{-5.5mm} \caption*{{\footnotesize Mask-ShadowGAN}} 
	\end{subfigure}
	\begin{subfigure}{0.16\textwidth}
		\includegraphics[width=\textwidth]{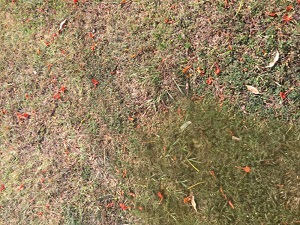}
		\vspace{-5.5mm} \caption*{{\footnotesize DSC-I~\cite{hu2019direction,wang2018stacked}}} 
	\end{subfigure}
	\begin{subfigure}{0.16\textwidth}
		\includegraphics[width=\textwidth]{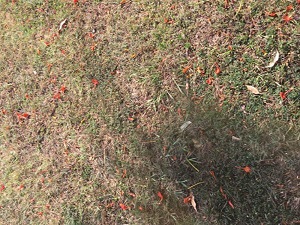}
		\vspace{-5.5mm} \caption*{{\footnotesize DSC-S~\cite{hu2019direction,qu2017deshadownet}}}
	\end{subfigure}
	\begin{subfigure}{0.16\textwidth}
		\includegraphics[width=\textwidth]{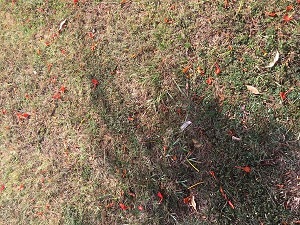}
		\vspace{-5.5mm} \caption*{{\footnotesize Gong~\etal~\cite{gong2014interactive}}}
	\end{subfigure}
	\begin{subfigure}{0.16\textwidth}
		\includegraphics[width=\textwidth]{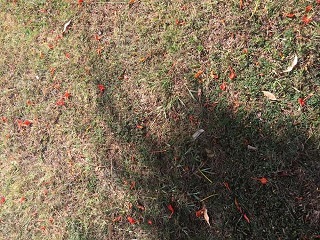}
		\vspace{-5.5mm} \caption*{{\footnotesize Guo~\etal~\cite{guo2013paired}}}
	\end{subfigure}

    \vspace{-3mm}	
	\caption{Comparing shadow removal results produced by various methods on the USR dataset.}
	%\caption{Visual comparison of shadow removal results produced by various methods on the USR dataset.}
	\label{fig:comparison_removal_USR}
	\vspace{-3mm}
\end{figure*}

%%%%%%%%%%%%%%%%%%%%%%%%%%%%%%%%%%%%%%%%%

%% file: figs/SRD_ISTD_comp.tex
%%%%%%%%%%%%%%%%%%%%%%%%%%%%%%%%%%%%%%%%%%%%%%%%%%%%%%%%%%%%%%%%%%%%
% teaser
%%%%%%%%%%%%%%%%%%%%%%%%%%%%%%%%%%%%%%%%%%%%%%%%%%%%%%%%%%%%%%%%%%%%

%%%%%%%%%%%%%%%%%%%%%%%%%%%%%%%%%%%%%%%%%

\begin{figure*}[tp]
	\centering
	\begin{subfigure}{0.138\textwidth} %0.138\textwidth; 0.195\textwidth 0.16\textwidth
\caption*{{\footnotesize input}}
\vspace*{-3mm}
	\end{subfigure}
	\begin{subfigure}{0.138\textwidth}
\caption*{{\footnotesize ground truth}} 
\vspace*{-3mm}
	\end{subfigure}
	\begin{subfigure}{0.138\textwidth}
\caption*{{\footnotesize Mask-ShadowGAN}} 
\vspace*{-3mm}
	\end{subfigure}
	\begin{subfigure}{0.138\textwidth}
\caption*{{\footnotesize DSC~\cite{hu2019direction}}} 
\vspace*{-3mm}
	\end{subfigure}
	\begin{subfigure}{0.138\textwidth}
\caption*{{\footnotesize ST-CGAN~\cite{wang2018stacked}}}
\vspace*{-3mm}
	\end{subfigure}
	\begin{subfigure}{0.138\textwidth}
\caption*{{\footnotesize Gong~\etal~\cite{gong2014interactive}}}
\vspace*{-3mm}
	\end{subfigure}
	\begin{subfigure}{0.138\textwidth}
\caption*{{\footnotesize Guo~\etal~\cite{guo2013paired}}}
\vspace*{-3mm}
	\end{subfigure}

	\vspace*{0.5mm}
	\begin{subfigure}{0.138\textwidth} %0.12\textwidth; 0.195\textwidth 0.16\textwidth
		\includegraphics[width=\textwidth]{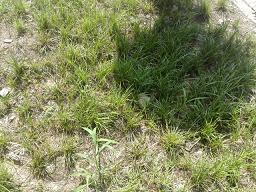}
	\end{subfigure}
	\begin{subfigure}{0.138\textwidth} %0.12\textwidth; 0.195\textwidth 0.16\textwidth
		\includegraphics[width=\textwidth]{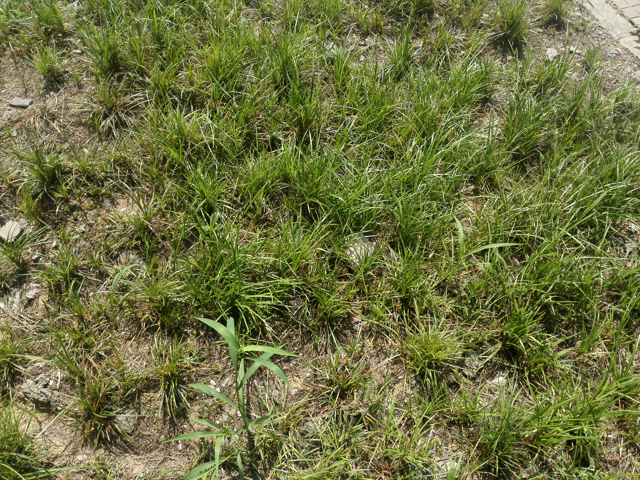}
	\end{subfigure}
	\begin{subfigure}{0.138\textwidth} %0.138\textwidth; 0.195\textwidth 0.16\textwidth
		\includegraphics[width=\textwidth]{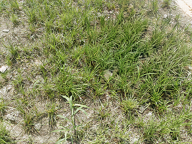}
	\end{subfigure}
	\begin{subfigure}{0.138\textwidth} %0.138\textwidth; 0.195\textwidth 0.16\textwidth
		\includegraphics[width=\textwidth]{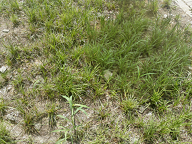}
	\end{subfigure}
	\begin{subfigure}{0.138\textwidth} %0.138\textwidth; 0.195\textwidth 0.16\textwidth
		\includegraphics[width=\textwidth]{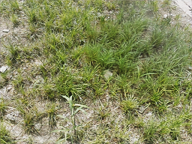}
	\end{subfigure}
	\begin{subfigure}{0.138\textwidth} %0.138\textwidth; 0.195\textwidth 0.16\textwidth
		\includegraphics[width=\textwidth]{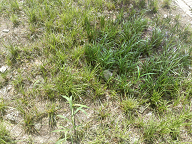}
	\end{subfigure}
	\begin{subfigure}{0.138\textwidth} %0.138\textwidth; 0.195\textwidth 0.16\textwidth
		\includegraphics[width=\textwidth]{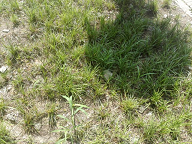}
	\end{subfigure}
	
	\vspace*{0.5mm}
	\begin{subfigure}{0.138\textwidth} %0.12\textwidth; 0.195\textwidth 0.16\textwidth
		\includegraphics[width=\textwidth]{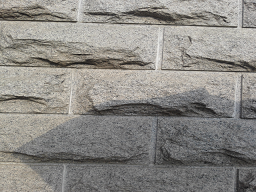}
	\end{subfigure}
	\begin{subfigure}{0.138\textwidth} %0.12\textwidth; 0.195\textwidth 0.16\textwidth
		\includegraphics[width=\textwidth]{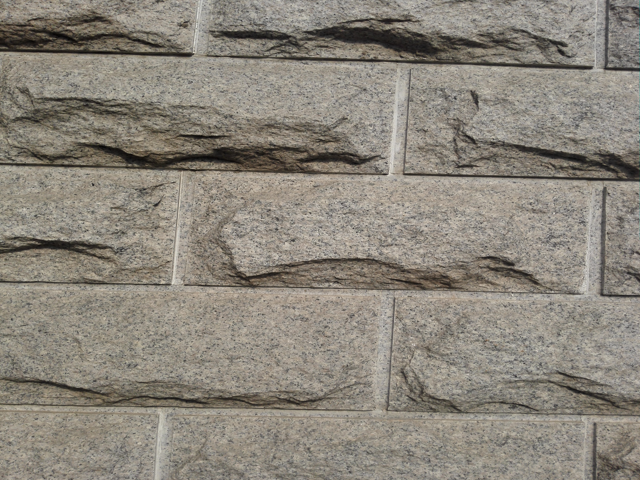}
	\end{subfigure}
	\begin{subfigure}{0.138\textwidth} %0.138\textwidth; 0.195\textwidth 0.16\textwidth
		\includegraphics[width=\textwidth]{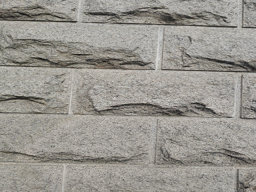}
	\end{subfigure}
	\begin{subfigure}{0.138\textwidth} %0.138\textwidth; 0.195\textwidth 0.16\textwidth
		\includegraphics[width=\textwidth]{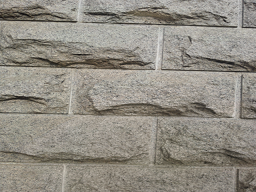}
	\end{subfigure}
	\begin{subfigure}{0.138\textwidth} %0.138\textwidth; 0.195\textwidth 0.16\textwidth
		\includegraphics[width=\textwidth]{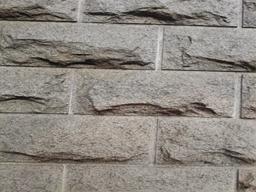}
	\end{subfigure}
	\begin{subfigure}{0.138\textwidth} %0.138\textwidth; 0.195\textwidth 0.16\textwidth
		\includegraphics[width=\textwidth]{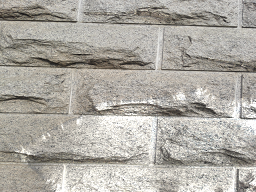}
	\end{subfigure}
	\begin{subfigure}{0.138\textwidth} %0.138\textwidth; 0.195\textwidth 0.16\textwidth
		\includegraphics[width=\textwidth]{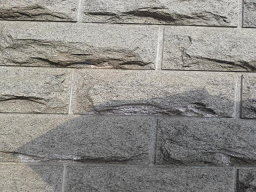}
	\end{subfigure}
    
    \vspace*{0.8mm}
    \begin{subfigure}{0.996\textwidth} %0.138\textwidth; 0.195\textwidth 0.16\textwidth
    	\includegraphics[width=\textwidth]{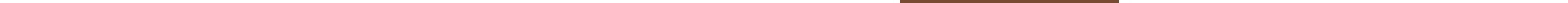}
    \end{subfigure}
	
	%\vspace*{0.2mm}
	\begin{subfigure}{0.138\textwidth} %0.12\textwidth; 0.195\textwidth 0.16\textwidth
		\includegraphics[width=\textwidth]{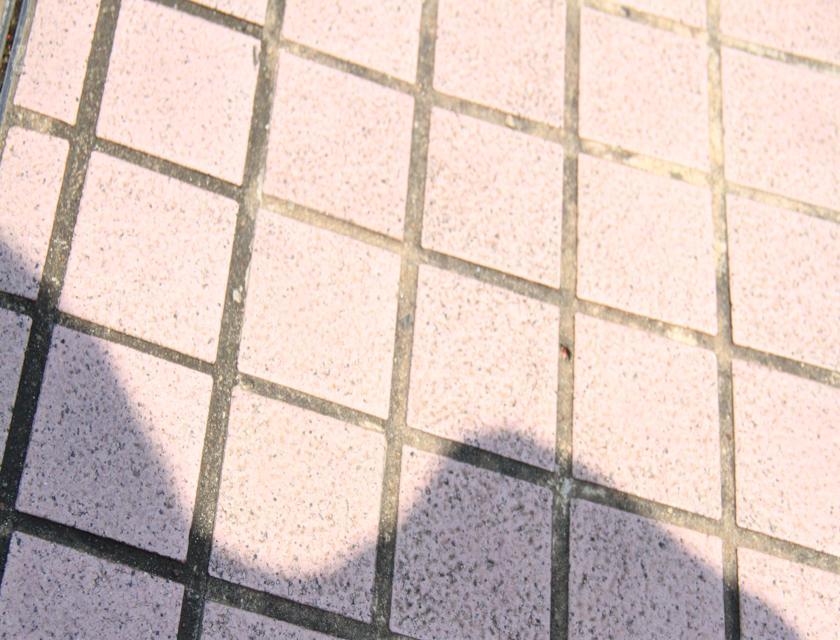}
	\end{subfigure}
	\begin{subfigure}{0.138\textwidth} %0.12\textwidth; 0.195\textwidth 0.16\textwidth
		\includegraphics[width=\textwidth]{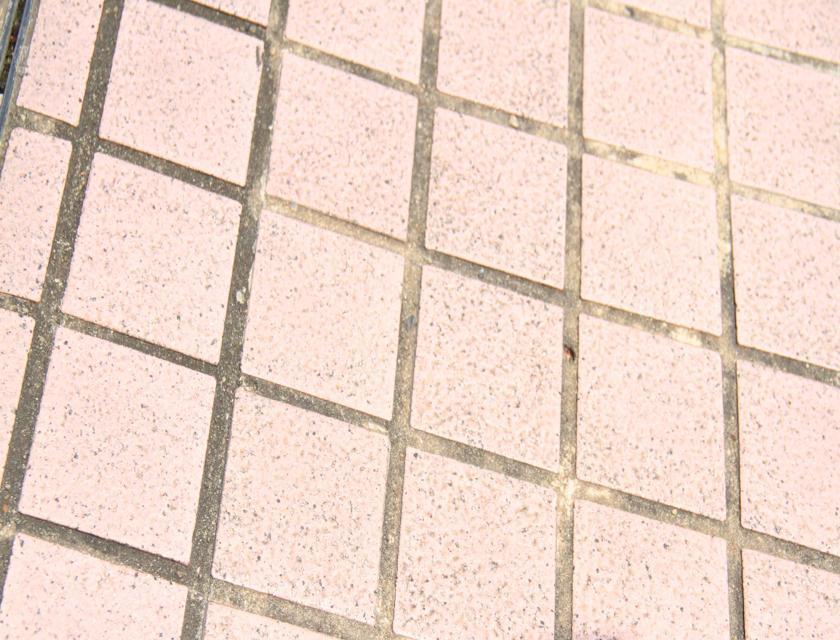}
	\end{subfigure}
	\begin{subfigure}{0.138\textwidth} %0.138\textwidth; 0.195\textwidth 0.16\textwidth
		\includegraphics[width=\textwidth]{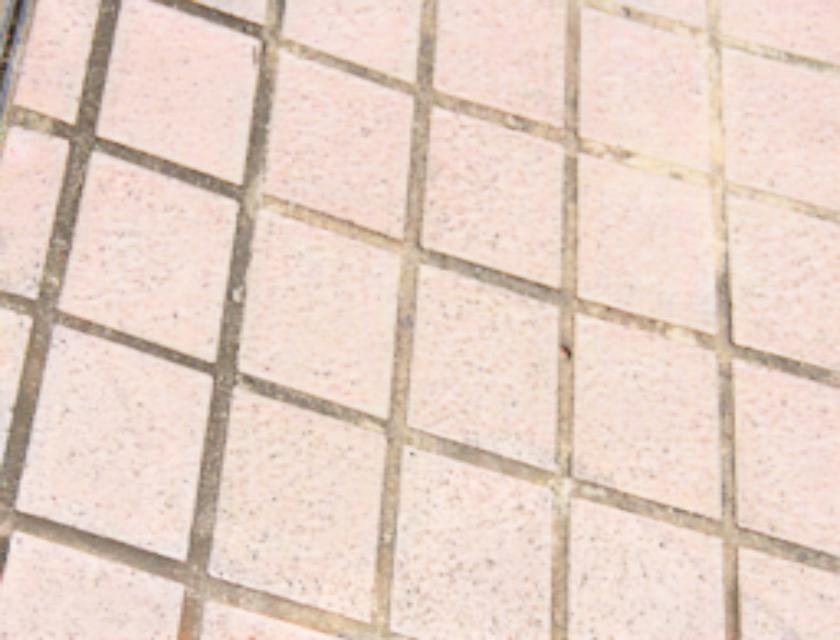}
	\end{subfigure}
	\begin{subfigure}{0.138\textwidth} %0.138\textwidth; 0.195\textwidth 0.16\textwidth
		\includegraphics[width=\textwidth]{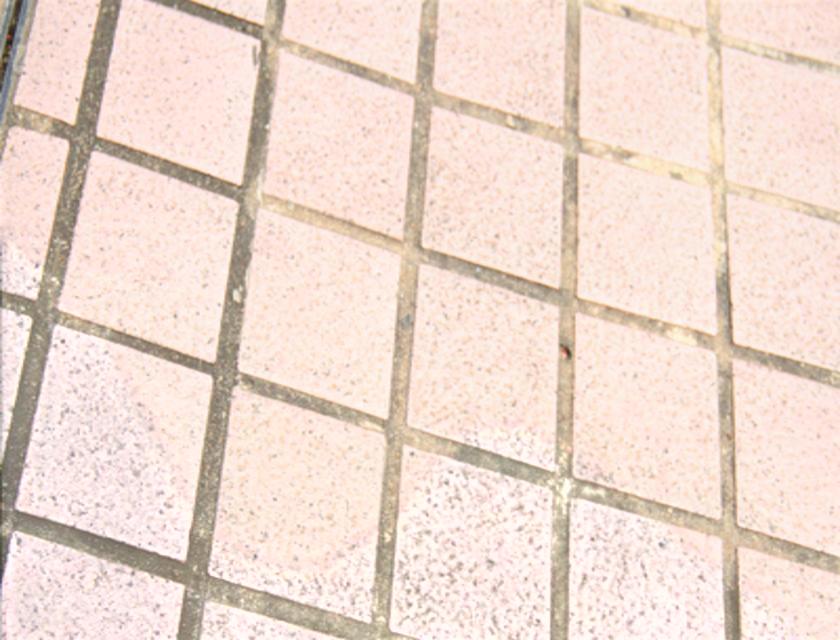}
	\end{subfigure}
	\begin{subfigure}{0.138\textwidth} %0.138\textwidth; 0.195\textwidth 0.16\textwidth
		\includegraphics[width=\textwidth]{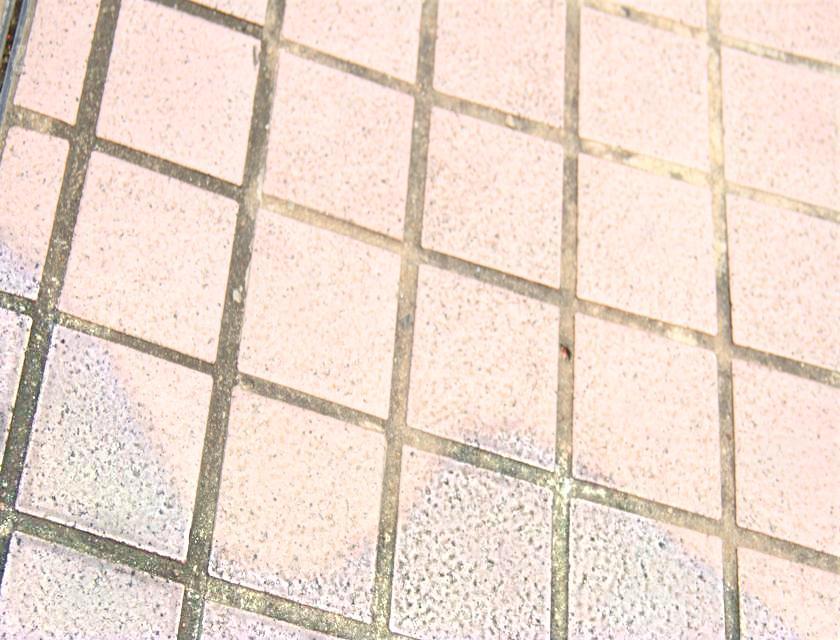}
	\end{subfigure}
	\begin{subfigure}{0.138\textwidth} %0.138\textwidth; 0.195\textwidth 0.16\textwidth
		\includegraphics[width=\textwidth]{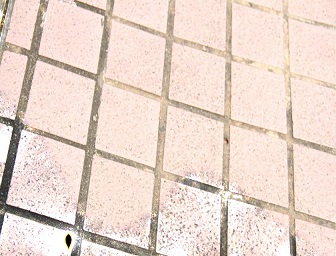}
	\end{subfigure}
	\begin{subfigure}{0.138\textwidth} %0.138\textwidth; 0.195\textwidth 0.16\textwidth
		\includegraphics[width=\textwidth]{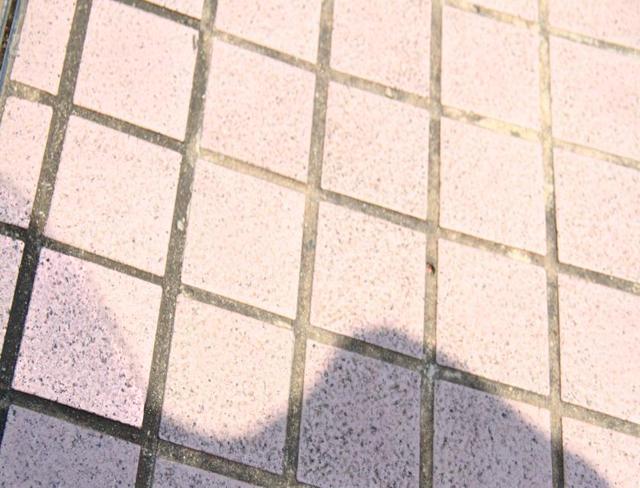}
	\end{subfigure}

	\vspace*{0.5mm}
	\begin{subfigure}{0.138\textwidth} %0.12\textwidth; 0.195\textwidth 0.16\textwidth
		\includegraphics[width=\textwidth]{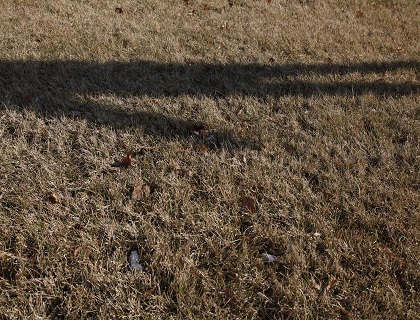}
	\end{subfigure}
	\begin{subfigure}{0.138\textwidth} %0.12\textwidth; 0.195\textwidth 0.16\textwidth
		\includegraphics[width=\textwidth]{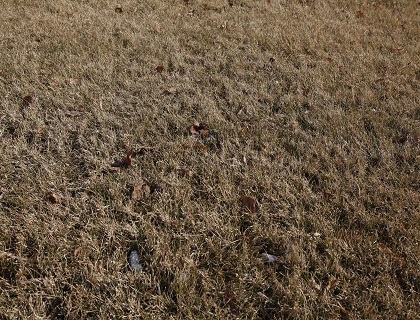}
	\end{subfigure}
	\begin{subfigure}{0.138\textwidth} %0.138\textwidth; 0.195\textwidth 0.16\textwidth
		\includegraphics[width=\textwidth]{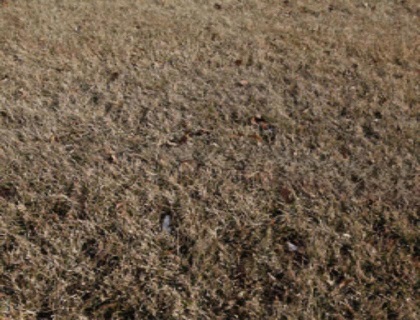}
	\end{subfigure}
	\begin{subfigure}{0.138\textwidth} %0.138\textwidth; 0.195\textwidth 0.16\textwidth
		\includegraphics[width=\textwidth]{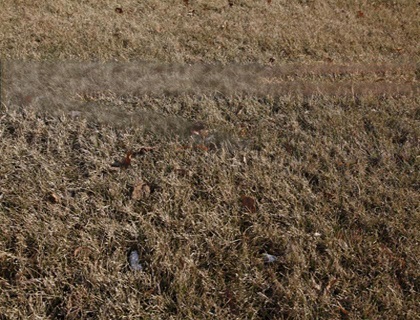}
	\end{subfigure}
	\begin{subfigure}{0.138\textwidth} %0.138\textwidth; 0.195\textwidth 0.16\textwidth
		\includegraphics[width=\textwidth]{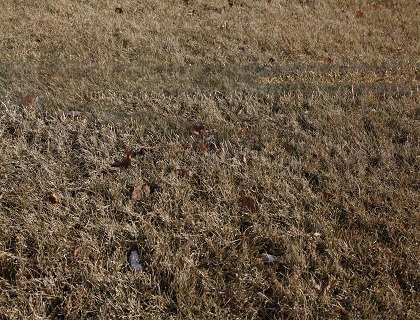}
	\end{subfigure}
	\begin{subfigure}{0.138\textwidth} %0.138\textwidth; 0.195\textwidth 0.16\textwidth
		\includegraphics[width=\textwidth]{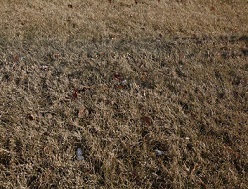}
	\end{subfigure}
	\begin{subfigure}{0.138\textwidth} %0.138\textwidth; 0.195\textwidth 0.16\textwidth
		\includegraphics[width=\textwidth]{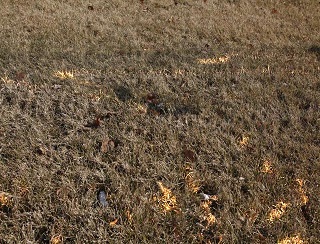}
	\end{subfigure}

	\vspace*{0.5mm}
	\begin{subfigure}{0.138\textwidth} %0.138\textwidth; 0.195\textwidth 0.16\textwidth
		\includegraphics[width=\textwidth]{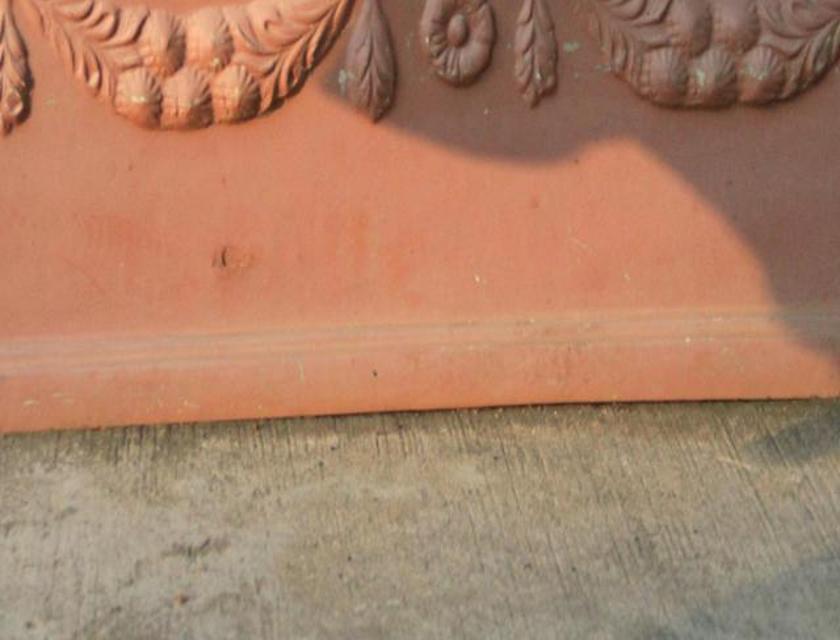}
		\vspace{-5.5mm} \caption*{{\footnotesize input}}
	\end{subfigure}
	\begin{subfigure}{0.138\textwidth}
		\includegraphics[width=\textwidth]{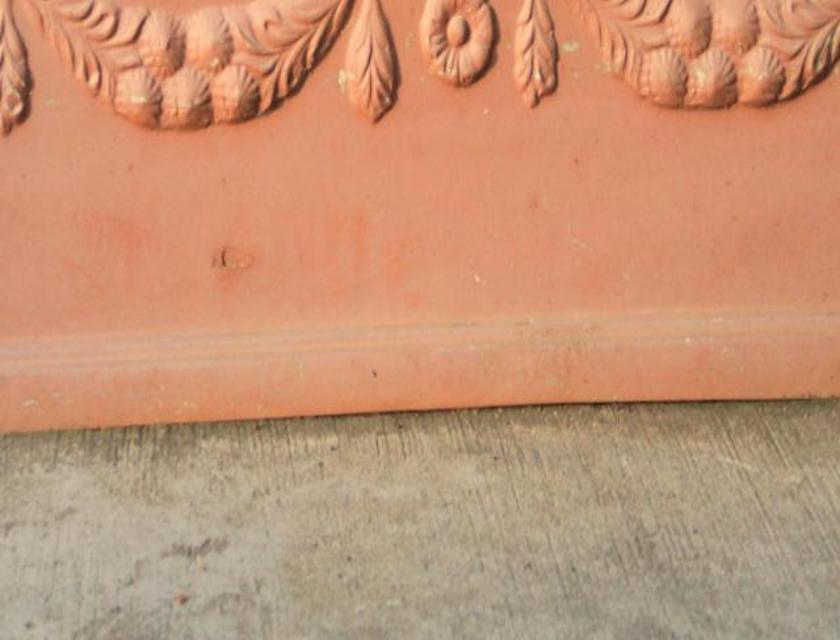}
		\vspace{-5.5mm} \caption*{{\footnotesize ground truth}} 
	\end{subfigure}
	\begin{subfigure}{0.138\textwidth}
		\includegraphics[width=\textwidth]{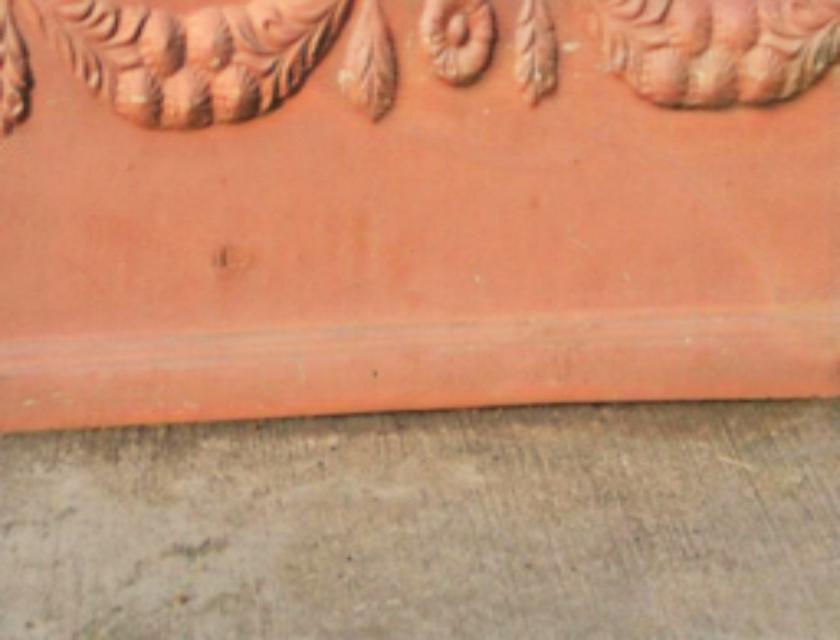}
		\vspace{-5.5mm} \caption*{{\footnotesize Mask-ShadowGAN}} 
	\end{subfigure}
	\begin{subfigure}{0.138\textwidth}
		\includegraphics[width=\textwidth]{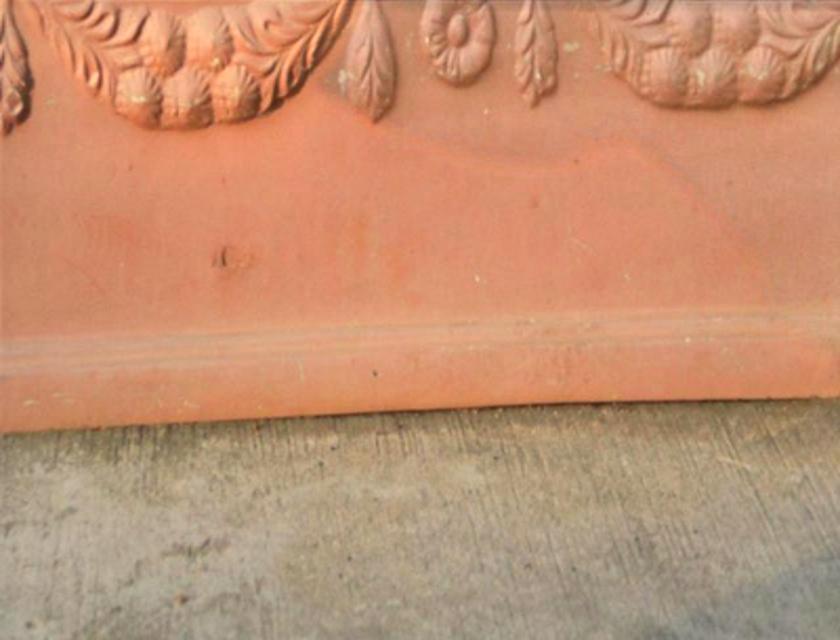}
		\vspace{-5.5mm} \caption*{{\footnotesize DSC~\cite{hu2019direction}}} 
	\end{subfigure}
	\begin{subfigure}{0.138\textwidth}
		\includegraphics[width=\textwidth]{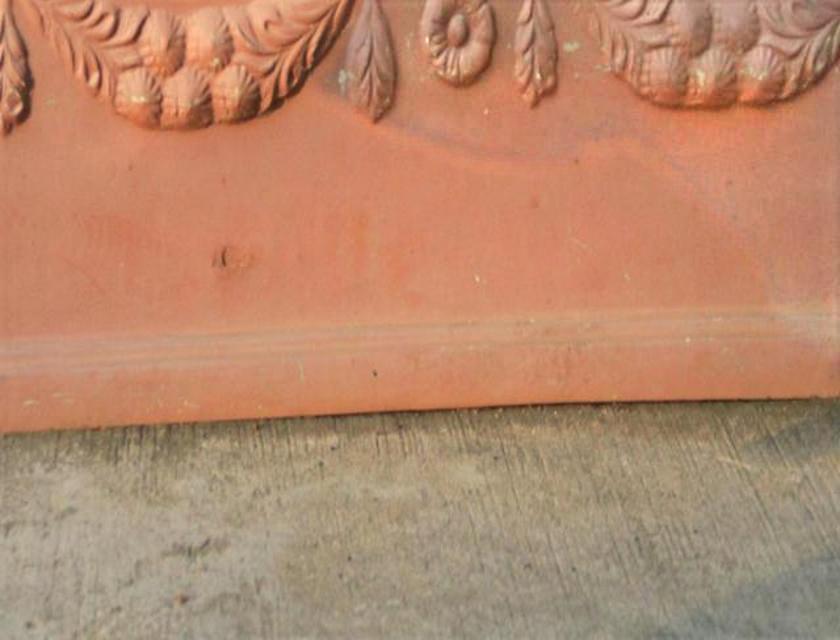}
		\vspace{-5.5mm} \caption*{{\footnotesize DeshadowNet~\cite{qu2017deshadownet}}}
	\end{subfigure}
	\begin{subfigure}{0.138\textwidth}
		\includegraphics[width=\textwidth]{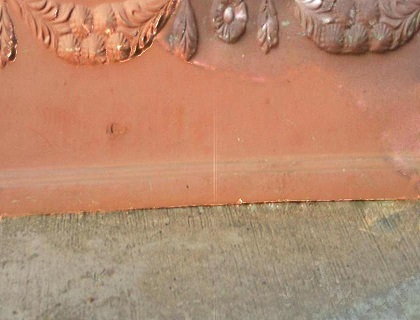}
		\vspace{-5.5mm} \caption*{{\footnotesize Gong~\etal~\cite{gong2014interactive}}}
	\end{subfigure}
	\begin{subfigure}{0.138\textwidth}
		\includegraphics[width=\textwidth]{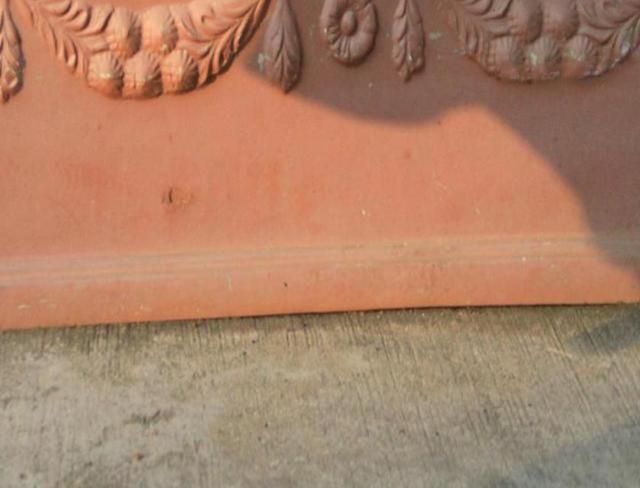}
		\vspace{-5.5mm} \caption*{{\footnotesize Guo~\etal~\cite{guo2013paired}}}
	\end{subfigure}

	\vspace{-3mm}
	\caption{Comparing shadow removal results on the ISTD~\cite{wang2018stacked} (top two rows) \& SRD~\cite{qu2017deshadownet} (last three rows) datasets. In the fifth column, the top two results are produced from ST-CGAN~\cite{wang2018stacked} while the others are produced from DeshadowNet~\cite{qu2017deshadownet}.}
	% Please zoom in for a better view.}
	%\caption{Visual comparison of shadow removal results on ISTD dataset~\cite{wang2018stacked} (the first two rows) and SRD dataset~\cite{qu2017deshadownet} (the last three rows). Note that in the fifth column, the top two results are produced from ST-CGAN~\cite{wang2018stacked} while the others are produced from DeshadowNet~\cite{qu2017deshadownet}. Please zoom in for a better view.}
	\label{fig:comparison_removal_SRD_ISTD}
	\vspace{-4mm}
\end{figure*}

%%%%%%%%%%%%%%%%%%%%%%%%%%%%%%%%%%%%%%%%%

%% file: tables/table_SRDISTD_comparison.tex
\begin{table}  [tp]
	\begin{center}
		\caption{Comparison with the state-of-the-art methods on the SRD~\cite{qu2017deshadownet} and ISTD~\cite{wang2018stacked} datasets in terms of RMSE. Note that the code of ST-CGAN~\cite{wang2018stacked} and DeshadowNet~\cite{qu2017deshadownet} is not publicly available, so we directly compare with their results on their respective datasets.}
		\vspace{-2mm}
		\label{table:SRDISTD_removal}
		\resizebox{1.0\linewidth}{!}{%
		\begin{tabular}{c|c|c|c}
			Training data & Methods&
			SRD~\cite{qu2017deshadownet} &
			ISTD~\cite{wang2018stacked}
			\\
			\hline
			\hline
			\multirow{2}{*}{unpaired} & \textbf{Mask-ShadowGAN} & \textbf{7.32} & \textbf{7.61} \\
			& CycleGAN~\cite{zhu2017unpaired} &  9.14 & 8.16 \\
			\hline
			\multirow{3}{*}{paired} & DSC~\cite{hu2019direction} & \textbf{6.21} &  \textbf{6.67}   \\
			
			&ST-CGAN~\cite{wang2018stacked} & - & 7.47\\
			
			&DeshadowNet~\cite{qu2017deshadownet} & 6.64 &  -    \\
			
			\hline
			
			\multirow{3}{*}{-}  &Gong \emph{et al.}~\cite{gong2014interactive} & 8.73 &  8.53  \\
			
			&Guo \emph{et al.}~\cite{guo2013paired} & 12.60 & 9.30 \\
			
			&Yang \emph{et al.}~\cite{yang2012shadow} & 22.57 & 15.63 \\
			\hline
			
		\end{tabular} }
		%}
	\end{center}
	\vspace{-7mm}
\end{table}

%% file: section6-conclusion.tex
\section{Conclusion}
\label{sec::conclusion}

In this work, we present a novel generative adversarial framework, named as Mask-ShadowGAN, for shadow removal based on unpaired shadow and shadow-free images.
%
%Our key idea is to learn the shadow mask from a real shadow image and leverage the mask as a guidance to indicate how to add shadows on the shadow-free images.
Our key idea is to transform the uncertain shadow-free-to-shadow image translation into a deterministic image translation
%,~\ie, many-to-one shadow removal and one-to-many shadow generation, 
with the guidance of the shadow masks, which are learned from the real shadow images automatically.
%
%Our key idea is to transform the uncertain shadow-free-to-shadow image translation into a deterministic and one-to-one image translation under the guidance of shadow masks, where the shadow masks are learned from the real shadow images automatically.
%
Further, we construct the first unpaired shadow removal (USR) dataset, test our method on various datasets, and compare it with the state-of-the-art methods to show its quality, both quantitatively and visually.
However, our method assumes a small domain difference (\eg, variations in scene contents) between the unpaired shadow and shadow-free images. 
Also, to aim for better results, we generally need a larger training set.
In the future, we plan to use the generated shadow masks to facilitate new applications, such that shadow editing, where we may manipulate shadows rather than just removing them.
We also plan to explore the mask generation technique for other image translation applications, where mapping between image sets is not one-to-one,~\eg, learn to remove/synthesize rain or snow from unpaired real rain/snow and rain-free/snow-free images. 
%In the future, we plan to collect more shadow and shadow-free images under more complex scenes, and enhance the capability of the network to remove shadows in videos.